\documentclass[journal]{IEEEtai}

\usepackage[colorlinks,urlcolor=blue,linkcolor=blue,citecolor=blue]{hyperref}

\usepackage{color,array}

\usepackage{graphicx}

\usepackage{graphicx}
\usepackage{amsmath,amsfonts}
\usepackage{algorithmic}
\usepackage{algorithm}
\usepackage{array}
\usepackage{textcomp}
\usepackage{stfloats}
\usepackage{url}
\usepackage{verbatim}
\usepackage{graphicx}
\usepackage{cite}
\usepackage{textcomp}
\usepackage{amssymb,multirow,setspace}
\usepackage{inconsolata}
\usepackage{bm}
\usepackage{booktabs} 
\usepackage{subfigure}
\usepackage{color}

 \usepackage{bbm}
\usepackage{amsmath,amsfonts}
\usepackage{algorithmic}
 \usepackage{array}
 \usepackage{multirow}
\usepackage{stfloats}
\usepackage{url}
\usepackage{cite}
\usepackage{verbatim}
\usepackage{balance}
\usepackage{pifont}

 \usepackage{bm}

\begin{document}

 \title{Adaptive Hierarchical Graph Cut for Multi-granularity Out-of-distribution Detection}

\author{Xiang Fang,~\IEEEmembership{Graduate Student Member,~IEEE}, Arvind Easwaran, Blaise Genest, Ponnuthurai Nagaratnam Suganthan,~\IEEEmembership{Fellow,~IEEE}%
\thanks{This research is part of the programme DesCartes and is supported by the National Research Foundation, Prime Minister’s Office, Singapore under its Campus for Research Excellence and Technological Enterprise (CREATE) programme. (Corresponding author: Xiang Fang)}
\thanks{Xiang Fang is with Energy Research Institute @ NTU, Interdisciplinary Graduate Programme, Nanyang Technological University, Singapore, is also with College of Computing and Data Science, Nanyang Technological University, Singapore, is also with CNRS@CREATE, Singapore (e-mail: xiang003@e.ntu.edu.sg).}
\thanks{Arvind Easwaran with College of Computing and Data Science, Nanyang Technological University, Singapore, is also with CNRS@CREATE, Singapore (e-mail: arvinde@ntu.edu.sg).}
\thanks{Blaise Genest is with CNRS and CNRS@CREATE, IPAL IRL 2955, France and Singapore (e-mail: blaise.genest@cnrsatcreate.sg).}
\thanks{Ponnuthurai Nagaratnam Suganthan is with KINDI Computing Research Center, College of Engineering, Qatar University, Doha (e-mail: p.n.suganthan@qu.edu.qa).}
}

\maketitle

\begin{abstract} 
This paper focuses on a significant yet challenging  task: out-of-distribution detection (OOD detection), which aims to distinguish and reject test samples with  semantic shifts, so as to prevent models trained on in-distribution (ID) data from producing unreliable predictions. Although previous   works have made decent success, they are ineffective for real-world challenging applications since these methods simply regard all unlabeled data as  OOD data and ignore the case that different datasets have different label granularity.
For example, ``cat'' on CIFAR-10 and ``tabby cat'' on Tiny-ImageNet share the same semantics  but have different labels due to various label granularity. To this end, in this paper, we propose a novel Adaptive Hierarchical Graph  Cut network (AHGC) to deeply explore the semantic relationship between different images. 
Specifically,  we construct a hierarchical KNN graph to evaluate the similarities  between different images based on the cosine similarity.
Based on the linkage and density information of the graph, we 
cut the graph into multiple subgraphs to integrate these semantics-similar samples. 
If the labeled percentage in a subgraph is larger than a threshold, we will assign the label with the highest percentage to unlabeled images.  
To further improve the model generalization, we augment each image into two augmentation versions, and  maximize the similarity between the two versions. 
Finally, we leverage the similarity score for OOD detection.
Extensive experiments on two challenging benchmarks (CIFAR-10 and CIFAR-100) illustrate that in representative cases, AHGC outperforms state-of-the-art OOD detection methods by 81.24\% on  CIFAR-100  and by 40.47\% on  CIFAR-10  in terms of ``FPR95'', which shows the effectiveness of our  AHGC. 
\end{abstract}

\begin{IEEEImpStatement}
As an effective method to detect outliers, out-of-distribution detection (OOD detection) has attracted more and more attention. However,  previous works simply regard all unlabeled data as  OOD data and ignore the case that some unlabeled samples might have the similar semantics with labeled data, which might render their methods ineffective for real-world challenging benchmarks. We construct a hierarchical KNN graph to evaluate the similarities  between different images.
Moreover, our method outperforms the state-of-the-art works by about 81.24\% in representative cases.
With satisfactory performance on multiple datasets, our method has wide potential applications.
\end{IEEEImpStatement}


\begin{figure}[t!]
\centering
\includegraphics[width=0.48\textwidth]{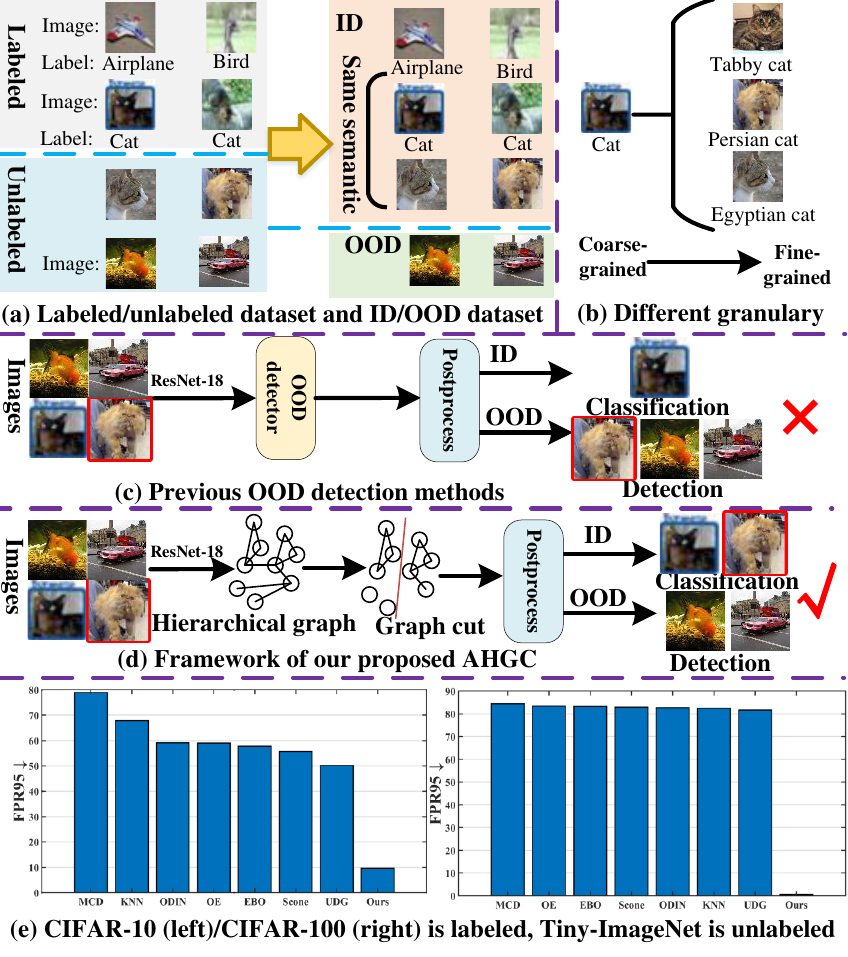}
\caption{(a) Relationship between labeled/unlabeled dataset and ID/OOD dataset. (b) Relationship between coarse- and fine-grained labels. In fact, ``Tabby cat'', ``Persian cat'' and ``Egyptian cat'' belong to the ``Cat'' class. (c) Previous methods ignore the relationship between  coarse- and fine-grained samples, and mistakenly treat unlabeled ID samples (Persian cat) as OOD.
(c) Brief framework of our proposed AHGC network,  detailed in Fig. \ref{fig:pipeline}. (d) Performance comparison (lower FPR95 value means  better performance) between  state-of-the-art methods and our AHGC on two semantically coherent out-of-distribution detection  benchmarks \cite{yang2021semantically}: the CIFAR-10 benchmark and the CIFAR-100 benchmark, where 
 left: CIFAR-10 \cite{krizhevsky2009learning} is used as the labeled dataset and 
Tiny-ImageNet \cite{le2015tiny} as the unlabeled dataset, 
and right: CIFAR-100 \cite{krizhevsky2009learning} is utilized as the labeled dataset and 
Tiny-ImageNet \cite{le2015tiny} as the unlabeled dataset.
Best viewed in color.}
\label{fig:intro}
\end{figure}

\section{Introduction}
Deep Neural Network (DNN) has achieved impressive 
success under a closed-set assumption \cite{arslan2021fine,yang2021garbagenet,gupta2020approaches,liu2020result}, where all the samples experienced during the test have been seen during training \cite{jiao2021multiscale,fang2020open,lu2023uncertainty}. 
In fact, 
standard DNN methods compulsorily classify every  sample as some of the known classes. The wrong classification of
outliers will result in irrecoverable losses in some safety-critical scenarios, such as autonomous driving \cite{zhang2021character,huang2022overview,cao2023accelerating}. To avoid these losses,  out-of-distribution (OOD) detection  \cite{hendrycksbaseline,karimi2022improving,de2023rule,shehu2021out,osman2022few,fang2021learning} is proposed to accurately detect the  outliers from OOD classes and correctly classify  the samples from in-distribution (ID) classes during testing.

The main challenge for OOD detection is that no  information about  OOD classes is available during training, making it difficult to distinguish ID and OOD samples. To  address the challenges, many  works \cite{liang2018enhancing,liu2020energy,sun2021react,lee2018simple,mohseni2020self,vyas2018out} calibrate the distribution of the softmax layer for OOD detection. Other methods \cite{yu2019unsupervised,zaeemzadeh2021out,hsu2020generalized,ming2023exploit} target to leverage a large number of OOD samples to learn the discrepancy between ID/OOD samples at  training time, then detect the OOD samples during testing. 
Although most OOD detection methods have achieved remarkable performance, they tend to perform unsatisfactorily  in  real-world artificial intelligence (AI) applications since they directly treat a labeled dataset as ID dataset while regarding all other datasets as OOD dataset. 
The treatment is inappropriate in many AI applications since an unlabeled dataset usually contains some ID images.
As shown in Fig. \ref{fig:intro}(a) and \ref{fig:intro}(b),   both the Tiny-ImageNet dataset and the CIFAR-10 dataset contain ``cat'' samples: The  Tiny-ImageNet dataset contains three fine-grained classes (``Tabby Cat'', ``Egyptian Cat'', and ``Persian Cat''), while the CIFAR-10 dataset only contains a coarse-grained class (``Cat''). Although these samples actually share the same semantics,  many methods \cite{chen2021adversarial,duvos,leetraining,sricharan2018building,vernekar2019out} regard them as different distributions in Fig. \ref{fig:intro}(c), which will severely limit their real-world performance since they ignore the relationship between labels with different granularity.
To overcome the above limitation, \cite{yang2021semantically} re-designs semantically coherent out-of-distribution detection (SC-OOD) benchmarks and proposes the UDG framework based on K-means clustering to dynamically filter ID samples from the unlabeled datasets.
Although \cite{yang2021semantically} finds there are some ID samples in unlabeled datasets,
the relationship between fine- and coarse-grained classes still
is overlooked. Since semantically coherent classes are treated as different classes, the dynamical K-means clustering often assigns wrong labels to unlabeled OOD samples, which results in unsatisfactory performance, as shown in Fig. \ref{fig:intro}(e).  There is still a large  room
to deal with OOD detection in real-world multi-granularity datasets.

To address the multi-granularity OOD detection task, we design a novel adaptive hierarchical graph cut network (AHGC) to explore the relationship between coarse-grained and fine-grained classes with the same semantics in Fig. \ref{fig:intro}(d).
Specifically, to mine the ID label information, based on the labeled dataset, we first train a feature encoder and a labeled classifier jointly for ID classification. Then, a hierarchical KNN graph is constructed to compute the cosine similarity between different images.
Besides, we leverage the ground-truth labels in the labeled dataset as supervision to compute the linkage and density of the graph. 
Based on the linkage and density information, we 
cut the graph into multiple subgraphs by removing low-weight edges from the graph. By the above graph cut operation, these high-similarity samples are clustered to the same subgraph, while low-similarity samples are assigned to  different subgraphs.
If the labeled percentage in a subgraph is larger than a predefined threshold, we will assign the label with the highest percentage to unlabeled images.  To enhance the image features, we re-extract the image feature after moving the unlabeled images into the labeled images.
To further improve the model generalization, we first map the enhanced features to a lower-dimensional logit space. Then, we augment each image into two augmentation versions, and  maximize the similarity between the two versions. 
Finally, given any unlabeled sample,
we utilize the energy-based approach with temperature-scaled logits  for OOD detection.

The main contributions of this paper are as follows:
\begin{itemize}
    \item We present a novel adaptive graph cut network, AHGC, which deeply explores  multi-granularity semantics between different   datasets. To the best of our knowledge, it is the first time that a graph cut network is proposed to detect  OOD samples and classify ID samples simultaneously.
    \item We propose an attention-aware graph cut approach to integrate the images across granularity. Besides,
    we design an iterative labeling strategy to encourage semantic alignment between different samples in the same subgraph in each epoch. Also, we augment each image into two versions and maximize their similarity for model generalization. 
    \item We conduct extensive experiments on two challenging  benchmarks (CIFAR-10 and CIFAR-100), where our proposed AHGC outperforms  state-of-the-art OOD detection methods with clear margins. In representative cases, AHGC outperforms all state-of-the-art OOD detection methods by 81.24\% on the CIFAR-100 benchmark and by 40.47\% on the CIFAR-10 benchmark in terms of ``FPR95'', demonstrating its effectiveness.
\end{itemize}

\section{Related Works}
\label{related}
\subsection{Out-of-distribution Detection}
Out-of-distribution detection  (OOD detection) targets to detect test samples from distributions that do not overlap with the training distribution \cite{liu2023exploring,wang2025taylor,fang2026towardsicml,kuai2026dynamic,wang2025point,fang2025your,zhang2025monoattack,fang2023hierarchical,liu2024towards,yang2025eood,fang2022multi,fang2026cogniVerse,lei2025exploring,fang2023you,wang2025dypolyseg,fang2025hierarchical,yan2026fit,fang2025adaptive,wang2026topadapter,cai2025imperceptible,fang2026slap,wang2026reasoning,fang2026immuno,wang2026biologically,fang2026disentangling,wang2025reducing,fang2026advancing,fang2026unveiling,wang2026from,liu2023conditional,liu2026attacking,fang2026rethinking,wang2025seeing,fang2026towards,fang2025multi,fang2024fewer,liu2024pandora,fang2024multi,fang2025turing,fang2024not,liu2023hypotheses,fang2024rethinking,liu2024unsupervised,fang2023annotations,xiong2024rethinking,fang2021unbalanced,wang2025prototype,zhang2025manipulating,fang2026align,tang2024reparameterization,fang2020double,tang2025simplification,fang2021animc,cai2026towards,fang2020v}. Previous OOD detection methods \cite{liang2018enhancing,liu2020energy,sun2021react,lee2018simple,mohseni2020self,vyas2018out,yu2019unsupervised,zaeemzadeh2021out,hsu2020generalized,ming2023exploit,kaur2022idecode,bibas2021single}  can be divided into four types: classification-based methods \cite{hendrycksbaseline,liang2018enhancing,lee2018hierarchical,leetraining}, density-based methods \cite{kirichenko2020normalizing,serrainput}, distance-based methods \cite{techapanurak2020hyperparameter,lee2018simple} and reconstruction-based methods \cite{zhou2022rethinking,yang2022out}. 
1) Early works mainly refer to a classification framework, which utilizes the maximum softmax probability to determine the ID/OOD samples. 
For example, by a two-step training process, \cite{mohseni2020self}  jointly learns generalizable OOD features and ID features with a minimal twist in a regular multi-class DNN classifier.
By using an auxiliary outlier dataset during training, POEM \cite{ming2022poem} selects informative outliers that are close to the decision boundary between ID and OOD samples.
2) To more explicitly model ID samples, density-based  methods  leverage the probabilistic models for OOD detection. These methods are based on an operating assumption that OOD samples have low likelihoods whereas ID samples have  high likelihoods under the estimated density model. {For instance,   ConjNorm \cite{pengconjnorm} introduce a novel Bregman divergence-based theoretical framework to design density functions in exponential family distributions.}
3) The distance-based  methods are under  an intuitive idea that OOD samples should be relatively far away from the centroids of ID samples. For example, \cite{lee2018simple} first utilizes a generative classifier under LDA assumption, and then use the minimum Mahalanobis distance to all class centroids for OOD detection.
4) The reconstruction-based methods often leverage the encoder-decoder framework, which is trained on only ID samples and generates different outcomes for OOD detection. For instance, MoodCat \cite{yang2022out}  is proposed to first mask a random image region and then determine the OOD samples based on  classification-based reconstruction results.

The above methods simply define one dataset as ID and all the others as OOD, which neglects the relevant semantic information between different datasets. To overcome the drawback, \cite{yang2021semantically}  re-designs these off-the-shelf popular benchmarks into two challenging and proper benchmarks, SC-OOD, and proposes a clustering-based network, UDG, for OOD detection.  Unfortunately, although UDG can separate some ID samples from unlabeled datasets, its clustering strategy cannot sufficiently understand the relationship between  datasets with different granularity, which causes it to not correctly assign correct labels to unlabeled ID samples. In this paper, we design a novel hierarchical graph cut framework to fully understand the relationship between different samples to widen the gap between ID and OOD samples.

\begin{figure*}[th!]
\centering
\includegraphics[width=\textwidth]{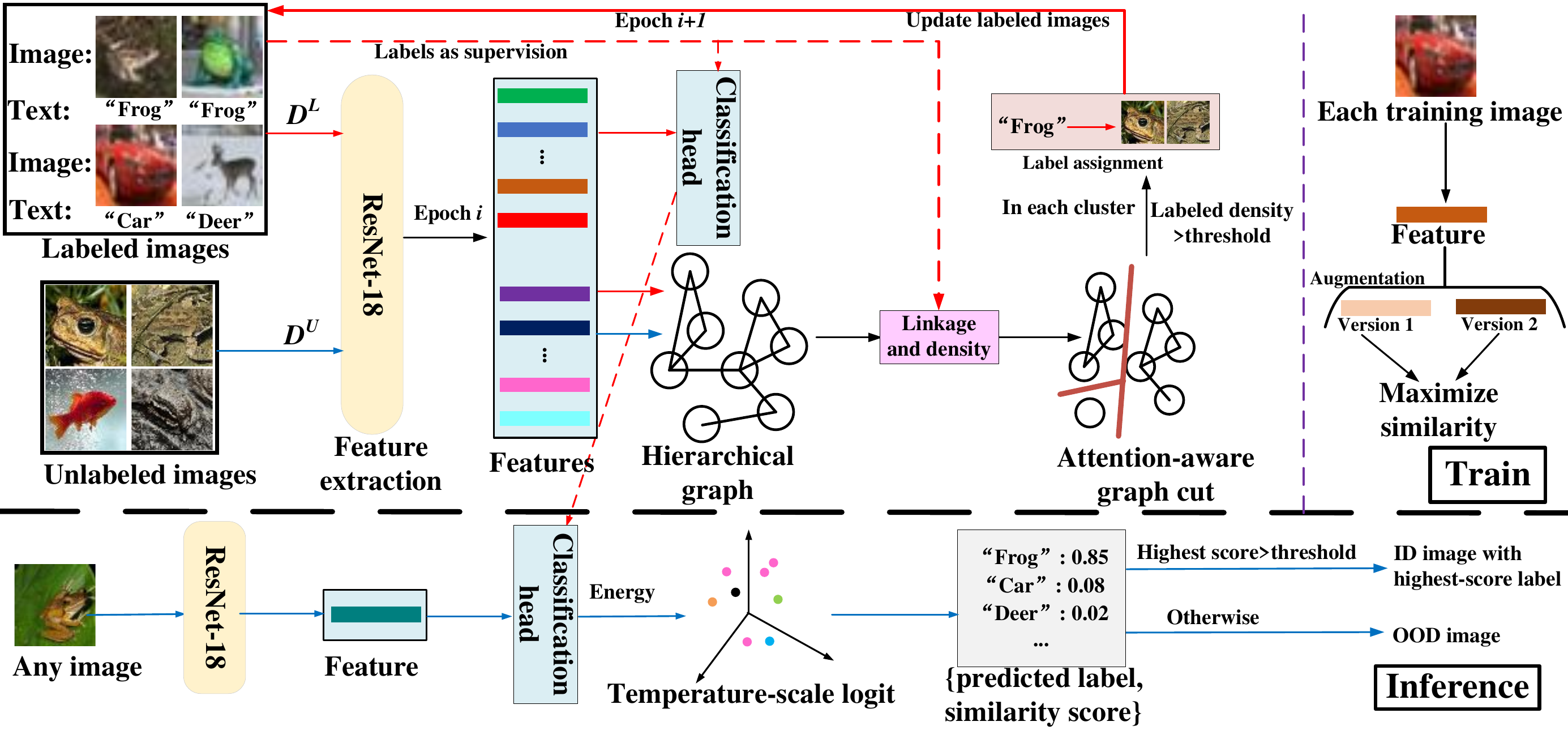}
\vspace{-10pt}
\caption{Overview of the proposed AHGC. Firstly, we utilize the ResNet-18 network \cite{he2016deep} to extract the visual features of labeled images $\mathcal{D}^L$ and unlabeled images $\mathcal{D}^U$. Then, we construct a hierarchical KNN graph to connect each image based on the cosine similarity. After that, we leverage the ground-truth labels in the labeled dataset as supervision to compute the linkage and density of the graph. Based on the linkage and density information, we 
cut the graph into multiple subgraphs by removing low-weight edges from the graph. If the labeled percentage is larger than a predefined threshold, we will assign the label with the highest percentage to unlabeled images. To improve the model generalization, we augment each image into two augmentation versions and then maximize the similarity between the two versions. During inference, with a classification head, we utilize a fully-connected layer to  map the image features into a logit space. Then, we leverage the similarity score from the logit as the evaluation of OOD samples. We color labeled data flow as {\textcolor{red}{red}} and unlabeled data flow as {\textcolor{blue}{blue}}.
Finally, we repeat these steps until the loss converges. Best viewed in color.
}
\vspace{-5pt}
\label{fig:pipeline}
\end{figure*}

\subsection{Graph Neural Networks} 
{As an important type of deep neural networks, 
graph neural networks (GNNs) \cite{ahmad2021graph,ragno2022prototype,wang2023multivariate,trappolini2023sparse,zou2023simple,yu2021recognizing,bianchi2020spectral,boykov2006graph} have been a topic of significant interest in the field of deep learning in recent years. Due to their satisfactory performance in many areas, GNNs have attracted more and more research attention. GNNs target extending  neural networks to deal with  graph-structured data based on the similarity between different samples.  
To process these graph-structured data, GNNs often iteratively update each node feature by aggregating feature similarity with its neighbor nodes. For example, 
GraphSAGE \cite{hamilton2017inductive} is introduced to aggregate feature representation from local neighborhoods. To aggregate and transform local information, the graph attention network \cite{velivckovicgraph} is designed to learn the weights of different neighbors during the node aggregation processes. Some GNN-based methods based on the graph cut strategy are proposed in the computer vision area. For instance,  a lightweight GNN is adopted in DeepCut \cite{aflalo2023deepcut}  with classical clustering objectives for unsupervised image segmentation. \cite{zhou2023attention} combines CNN and GCN to construct a multiple feature fusion model for hyperspectral image classification. Previous GNN-based methods refer to the closed-set assumption, and cannot be used in the challenging OOD detection task. 
Since GNNs can effectively evaluate the similarity between samples, we design a novel adaptive hierarchical graph cut  network to distinguish ID and OOD samples for OOD detection.}

\section{Problem Statement and Notations}\label{problem_state}

The OOD detection task aims to detect  outliers from the intended data
distribution,  to prevent models trained on ID samples from producing arbitrary predictions for OOD samples. 
For a benchmark with  labeled  and  unlabeled datasets, we denote the labeled dataset as $\mathcal{D}^L$
and unlabeled dataset as $\mathcal{D}^U$, respectively\footnote{In this paper,  superscripts $L$ and $U$ denote labeled and unlabeled samples, respectively;  superscripts $I$ and $O$ denote ID and OOD samples, respectively.}.
Most previous works \cite{liang2018enhancing,liu2020energy,sun2021react,lee2018simple,mohseni2020self,vyas2018out} assume that all the unlabeled samples in  $\mathcal{D}^U$ are OOD samples, while all the ID samples are in the  labeled dataset. 
However, real-world  unlabeled dataset $\mathcal{D}^U$ is a mixture of ID samples and OOD samples, \textit{i.e.}, some  unlabeled samples in $\mathcal{D}^U$ share the same semantics with labeled samples in $\mathcal{D}^L$, while the others have different semantics. Thus, we aim to mine the deep semantic features of labeled and unlabeled datasets.
On the challenging SC-OOD benchmarks, the training set $\mathcal{D}$ consists of labeled training set $\mathcal{D}^L$ and unlabeled training set $\mathcal{D}^U$. 
For the unlabeled dataset $\mathcal{D}^U$ including ID set $\mathcal{I}$ and OOD set $\mathcal{O}$, we target to widen their distance for detecting OOD set $\mathcal{O}$.

All the labeled samples  in $\mathcal{D}^L$ belong to $\mathcal{I}$, while the $\mathcal{D}^U$ is a mixture of partial ID set $\mathcal{D}_I^U$ from $\mathcal{I}$ and OOD set $\mathcal{D}_O^U$ from $\mathcal{O}$. 
Therefore, the training set is represented as $\mathcal{D}=\mathcal{D}^L\cup\mathcal{D}_I^U\cup \mathcal{D}_O^U$, where $\mathcal{D}^L\subset \mathcal{I}$, $\mathcal{D}_I^U\subset \mathcal{I}$ and $\mathcal{D}_O^U\subset \mathcal{O}$. 
Similarly, for the test set, we have $\mathcal{T}=\mathcal{T}_I \cup \mathcal{T}_O$, where $\mathcal{T}_I\subset \mathcal{I}$ and $\mathcal{T}_O\subset \mathcal{O}$.

For the labeled dataset $\mathcal{D}^L=(X^L, Y^L)$ with $N^L$ samples from $R$ classes, we have $X^L=(x_i^L)_{i=1}^{N^L}$ and labels $Y^L=(y_i^L)_{i=1}^{N^L}$, where $x_i$
denotes the $i$-th labeled sample and $y_i^L\in\{1,2,\dots,R\}$ is its corresponding label. Also, we collect an unlabeled dataset $\mathcal{D}^U=(X^U)$ with $N^U$ samples $X^U=(x_j^U)_{j=1}^{N^U}$. A carefully-designed model should not only accurately detect OOD samples from $\mathcal{D}^U$, but also correctly classify ID samples from $\mathcal{D}^L$ and $\mathcal{D}^U$ into $R$ classes.

\section{Adaptive Hierarchical Graph Cut Network}\label{method}
{We present our proposed AHGC network in Fig. \ref{fig:pipeline}, where we cluster labeled and unlabeled images into multiple subgraphs  and assign proper labels to unlabeled ID images based on density and linkage within each subgraph.
Firstly, we  feed all the images into the feature encoder to extract their features. To grasp the ID information from labeled images,
we co-train the feature encoder and a classification head. Then, we design a hierarchical graph network to integrate multi-granularity images. We construct the graph to model these images and cut low-similarity edges based on the density and linkage to obtain multiple subgraphs.
Based on the labeled percentage in each subgraph, we assign the proper labels to unlabeled images. Also, we integrate these cross-granularity images by subgraph aggregation.  Finally, we augment each image into two versions for better model generalization.}

\subsection{Feature Encoder and Classifier on Labeled Datasets}

Following \cite{yang2021semantically}, we use ResNet-18 \cite{he2016deep}  to extract  image features. 
For any sample $x_i^L$ (or $x_j^U$), we denote its feature  as $f_i^L\in\mathbb{R}^{d\times 1}$ (or $f_j^U\in\mathbb{R}^{d\times 1}$), 
where $d$ denotes the feature dimension.
Since only the labeled datasets contain the known label information of ID samples, we design a classifier to learn the label information for the ID classification task. Specifically, we target to train the  feature encoder $\mathcal{F}$ and a classification head $\mathcal{C}^L$ simultaneously. Given the image-label pairs ($X^L,Y^L$), we train the ID classifier with the feature encoder by the following loss: 
\begin{align}
    \mathcal{L}_0^L=\mathcal{L}_{CE}(X^L,Y^L,\theta_F,\theta_C),
\end{align}
where $\theta_F$ is a learnable parameter in $\mathcal{F}$, $\theta_C$ is a learnable parameter in $\mathcal{C}^L$, and 
$\mathcal{L}_{CE}(\cdot)$ denotes the standard cross-entropy loss. Based on $\mathcal{L}_0^L$, we can jointly train the feature encoder and the  classifier for labeled images. 
After training the classifier, we can understand the relationship between labels and samples for the following process.

\subsection{Single-level Graph Cut Mechanism for Base Model} 
\label{single_graph_cut}
{To measure the similarity between different images, we first construct a  graph by combining all images (both labeled samples $f_{i}^L$ and unlabeled images $f_{j}^U$). 
By jointing labeled image features and unlabeled  image features,
we can obtain the feature set $F = \{f_{i}^L\}_{i=1}^{N^L} \cup \{f_{j}^U\}_{j=1}^{N^U}=\{f_{i}\}_{i=1}^{N}$, where $N=N^L+N^U$ denotes the total image number.
Then, based on the cosine similarity,
we seek $K$-nearest neighbors of each image to construct an affinity graph $G = \{V, E\}$, where $V$ is the vertex set containing all images ({\em i.e.},  $|V|=N$), and $E$ is the edge set. In real-world  benchmarks (\textit{e.g.}, SC-OOD), each image entails one core object, which is represented as a node in the constructed graph $G$, with the corresponding image feature as the node feature. Each node is connected to its $k$ neighbors by the edges. }

We introduce a function $\vartheta(\cdot,\cdot)$ to construct the  edge subset $E^\prime \subset E$, where $E^\prime = \vartheta(G, F)$. The graph $G$ can be updated as $G'=\{V, E^\prime\}$, which is divided into multiple subgraphs, and each subgraph corresponds to a cluster of nodes. By the above graph cut paradigm, we design the KNN graph cut approach.

\subsection{Hierarchical Framework for Multi-granularity Graph Cut}
{Real-world AI applications adopt different labeling criteria, and different image datasets often offer various labels from a variety of granularity perspectives. For example, on the CIFAR-10 benchmark in Fig. \ref{fig:intro}(b), the cat only contains a label (``Cat''), while  in the Tiny-ImageNet dataset, the cat is divided into three labels (``Tabby Cat'', ``Egyptian Cat'', and ``Persian Cat''). Thus, we should deeply understand each image and explore the similarity between images.
Unfortunately, previous OOD detection methods \cite{chen2021adversarial,duvos,leetraining,sricharan2018building,vernekar2019out,yang2021semantically} ignore this multi-granularity situation, which severely degrades their performance in real-world AI applications. 
To solve the  problem, we propose a hierarchical framework based on the above single-level  graph cut paradigm in Section \ref{single_graph_cut}.}

Given initial image features $F=\{f_i\}_{i=1}^N$ and a small $k$, we design a hierarchical graph framework using a base subgraph function $\vartheta(\cdot,\cdot)$ and an aggregation function $\delta(\cdot,\cdot)$. Especially, we iteratively generate of a sequence of graphs $G_l = \{V_l, E_l\}$ and the corresponding node features $F_{l} = \{f_i\}_{i=1}^{|V_l|}$, where the subscript $l$ denote the $l$-th level and we  determine it as needed.

Firstly, we initialize the graph as $G_1$ and the corresponding node feature as $F_1 = \{f_i\}_{i=1}^{|V_1|}$. At level $l$, by introducing the subgraph function $\vartheta(\cdot,\cdot)$, we can obtain the initial edge subset: $E^\prime_{l} = \vartheta(G_l, F_l)$.
By introducing the function $\vartheta(\dot,\dot)$, we can obtain the selected edge subsets $E^\prime_{l}$ based on the node features $H_l$ and $k$-NN graph $G_l$. 

Then, we reserve the selected edge subsets $E^\prime_{l}$ and remove the rest edges. The graph $G_l$ can be updated as $G_l^\prime = \{V_l, E_l^\prime\}$. Therefore, we can split $G_l^\prime$ into multiple  subgraphs $\{c^{(l)}_i\}_{i=1}^{|V_{l+1}|}$, where $c^{(l)}_i$ is the $i$-th subgraph.
To obtain $G_{l+1}$ at the $(l+1)$-th level, we denote the $i$-th node as  $v_i^{(l+1)}$, which is an entity representing the subgraph $c_i^{(l)}$. 

After that, we introduce an aggregation function $\delta(\cdot,\cdot)$ to update the new node feature vectors: $F_{l+1} = \delta(F_l, G^\prime_l)$.
Therefore, we can  aggregate the node features in each subgraph $c_i^{(l)}$ into a single feature vector. 

Finally, we can search $k$-nearest-neighbors on $F_{l+1}$ and connect each node to its $k$ neighbors to obtain the updated $E_{l+1}$.
We repeat the above process until no more new edges are added, {\em i.e.}, $E_l^\prime = \varnothing$.  We denote  the length of the converged sequence as $Q$. 
Based on $G_Q$, we assign subgraph index $i$ to the subgraph $c^{(Q)}_i$, which propagates the index $i$ to all its nodes $\{v^{(Q)}_j|v^{(Q)}_j\in c^{(Q)}_i\}$. Then, each node $v^{(Q)}_i$ propagates its label to the subgraph $c^{(Q-1)}_i$ on the previous iteration.  By the above  process, we can assign a subgraph ID to each node in $V_1$, and finish the label assignment on the final level.

In the next sections, we will present the details about the base subgraph function $\vartheta(\cdot,\cdot)$ and the aggregation function $\delta(\cdot,\cdot)$. Besides, we can use our AHGC for underlying single-level model, akin to a single iteration of AHGC. Also, the  inference process will be described.

\begin{figure*}[t!]
\centering
\includegraphics[width=\textwidth]{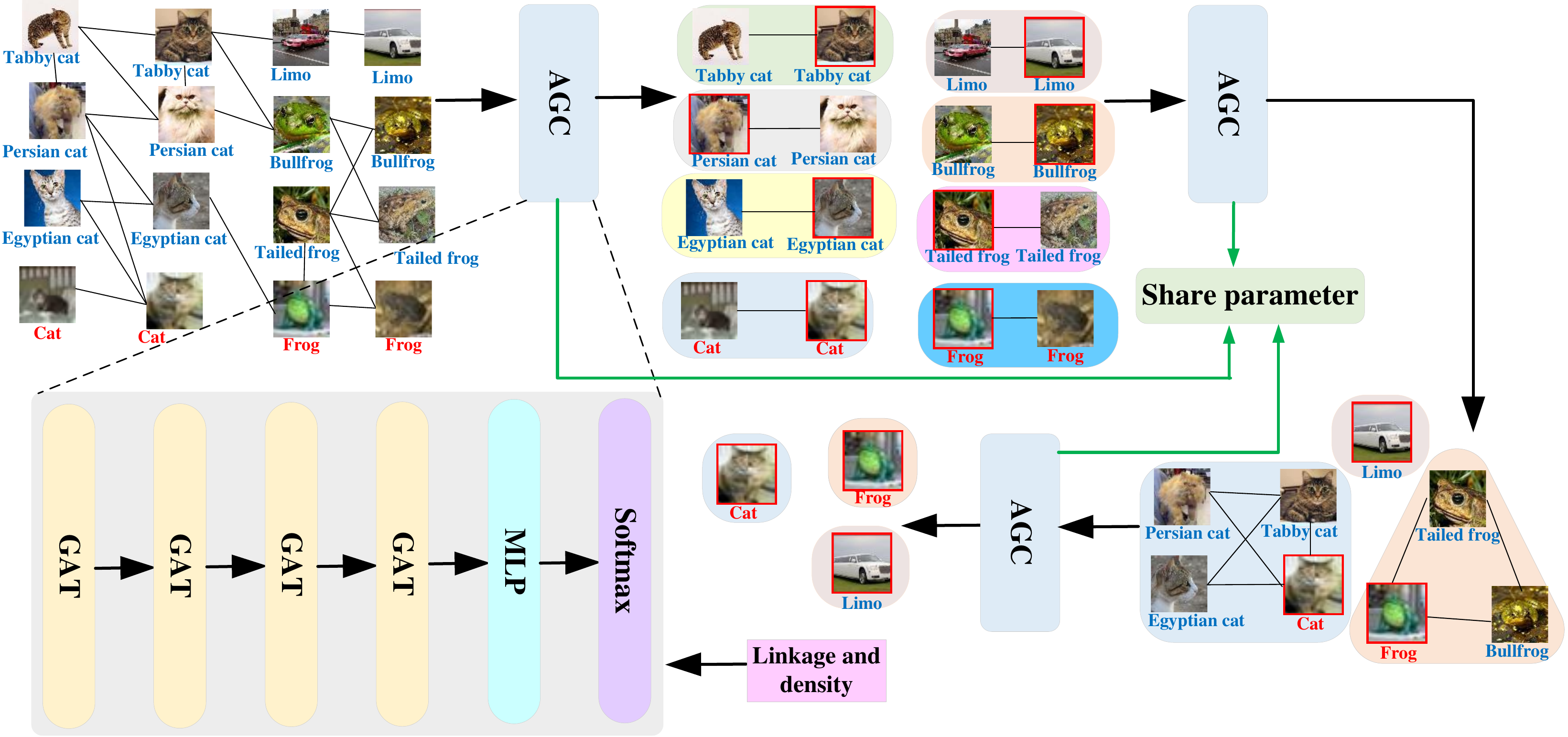}
\caption{Illustration of the attention-aware  graph cut  module, where ``AGC'' means the ``adaptive graph cut'' module,   the image with a \textcolor{red}{red} edge in a subgraph is its peak node, and labels are attached to labeled images. Multiple levels of AGC are utilized into the framework, where each level contains four GAT layers and an MLP layer followed by a softmax operation. Based on each AGC, we can divide the large graph into multiple subgraphs, and choose the representative node as the peak node. Therefore, we can integrate the multi-granularity images with the same semantics into a subgraph.  Best viewed in color.}
\label{multi-level}
\vspace{-5pt}
\end{figure*}

\subsection{Attention-aware Graph Cut via  Linkage and Density}
{Since the linkage and density information in the GNN model \cite{singh2023supervised,xing2021learning,zou2022improving} contain the rich  relationship between  different images, we treat them as supervision for model training.
To understand the semantics of images, we  use the graph attention network (GAT) \cite{velivckovic2018graph}  as the backbone to model the similarity between images by the linkage and density information. 
To extract the complex subgraph structures, we first utilize the graph cut function $\vartheta(\cdot,\cdot)$ for building a learnable graph network, where each node $v_i$ in $V$ comes with a subgraph label $y_i$. Unlike previous unsupervised graph cut methods \cite{fu2015normalized,khan2019graph} that focus on an explicit graph cut, we design the attention-aware  graph cut module based on data. 
To conduct the graph cut efficiently, we leverage a single graph encoder for embeddings to jointly predict these two quantities. Then, we utilize a  graph decoding process to estimate the linkage and density for determining edge connectivity and conducting graph cut.
Therefore, we detail the design of our unified joint linkage and density estimation model for graph cut. With the above considerations, we jointly predict the graph edge linkage and node density information based on a single full-graph inference. The subgraph prediction is passed through a graph decoding step.}

\noindent\textbf{Linkage and density for graph encoding.}
To effectively integrate the semantics-similar samples, we explore the relationship between different samples based on linkage and density.
For each node  feature $f_i$, we utilize
a stack of GAT network layers to encode it as the  embedding $f^\prime_i$. 
For each edge $(v_i, v_j)$ in the edge set $E$, we leverage the graph to concatenate the source and destination node features ($[f^\prime_i, f^\prime_j]$), where $i\neq j$ and $[\cdot, \cdot]$ denotes the concatenation operator. Then, we feed the concatenated features into a Multi-Layer Perceptron (MLP) layer \cite{rosenblatt1962principles} followed by a softmax transformation to produce the linkage probabilities: $p_{ij} = P(y_i = y_j)$,
where $p_{ij}$ is the estimate of the probability that two nodes linked by the edge share the same label.
Besides, we introduce a density proxy $d_{i}$ to measure the similarity-weighted proportion of same-subgraph nodes in its neighborhood for density estimation.

For any two nodes $v_i$ and $v_j$, we first compute their similarity $s_{ij}$ as the inner product of their respective  node features: $s_{ij}=\langle f_i, f_j\rangle$.
Then, we can obtain the corresponding edge coefficients as ${e}_{ij} = P(y_i = y_j) - P(y_i \neq y_j)$,
where $j$ denotes the $j$-th  neighbor of $v_i$. To approximate the ground-truth density $\bar{d}_i$, we introduce the estimator ${d}_{i}$
as follows:
\begin{equation} \label{eq:approximate_density}
    {d}_{i}  = \frac{1}{k}\sum\nolimits_{j=1}^{k} {e}_{ij} \cdot a_{ij}.
\end{equation}
By Eq.~\eqref{eq:approximate_density}, 
${d}_{i}$ can be obtained by  updating ${e}_{ij}$ with $e_{ij} = \mathbbm{1}(y_i = y_j) - \mathbbm{1}(y_i \neq y_j)$ using the ground-truth class labels, where $\mathbbm{1}$ is the indicator function.  
An ideal density $d_i$ is large if the most similar neighbors share the same labels; otherwise, it is small. 
By approximating $\bar{d}_i$ based on ${e}_{ij}$ and $p_{ij}$, the  joint  mechanism can reduce parameters for the prediction head during training.

\noindent\textbf{Edge selection for graph decoding.}  
During graph decoding, we convert them into final subgraphs by graph decoding. Especially, we tailor the graph decoding to incorporate our joint density and linkage estimates. Initially, we start with $E^\prime = \varnothing$. Given ${e}_{ij}$, ${d}_{i}$, $p_{ij}$ and an edge connection threshold $p_\tau$,  we first define a candidate edge set $\mathcal{Z}(i)$  for node $v_i$ as
\begin{equation}
\label{eq:e_def}
    \mathcal{Z}(i) = \{j|(v_i, v_j)\in E, p_{ij} \geq p_\tau  \text{ and } {d}_{i}\leq {d}_{j}\}.
\end{equation}
By the condition ${d}_i \le {d}_j$, we introduce an inductive bias to establish connections. 
For any $i$, if its candidate set $\mathcal{Z}(i)$ is not empty, we pick $j = \operatorname*{argmax}_{j\in\mathcal{Z}(i)} {e}_{ij}$, and then add ${e}_{ij}$ to $E^\prime$.
Based on Eq. \eqref{eq:e_def},  each node $v_i$ with a non-empty $\mathcal{Z}(i)$ can add exactly one edge to $E^\prime$. On the contrary, each node with an empty $\mathcal{Z}(i)$ becomes a peak node with no outgoing edges. 
The nodes with low density tend to 1) have some neighborhoods that overlap with other classes, 2) be near the boundary among different classes, 3) contain undesirable connections to other nodes. 
After a full pass over each node, $E^\prime$ forms a set of subgraphs $G^\prime$, which serve as the designated subgraphs.

\subsection{Intra-subgraph Label Assignment}
{After the graph cut, the unlabeled ID images and the labeled ID images will be clustered into the same subgraph.
Therefore, we aim at assigning labels to unlabeled ID images based on the labeled ID images.
After obtaining the subgraph indexes of all the samples $\mathcal{C} = \{c_{ij}\}_{i=1,j=1}^{i=N,j=K}$, we assign pseudo labels to unlabeled ID images based on the proportion of each label in a subgraph. At the $t$-th epoch, we denote the $k$-th subgraph as $\mathcal{D}_k^{(t)}=\{x_i| c^{(t)}_{ij}=k\}$, where $x_i$ is its $i$-th sample.
The set of ID-class labels is denoted as  $\mathcal{Y}^{(t)} = \mathcal{Y}^L \cup \mathcal{Y}_{gc}^{(t)} \in \{0,1,\dots,R-1\}$, where $R$ denotes the number of ID classes,  $\mathcal{Y}^L$ is formed by the ground-truth labels from $\mathcal{D}^L$, and $\mathcal{Y}_{gc}^{(t)}$ denotes the ID pseudo labels assigned to unlabeled images. When $t=0$, $\mathcal{Y}_{gc}^{(t)} = \varnothing$, while the ID pseudo labels will be added into $\mathcal{Y}_{gc}^{(t)}$ during training. Then, we define ``labeled percentage'' as the ratio of images belonging to class $y \in \mathcal{Y}^{(t)}$ in subgraph $k$:  $r_{k,y}^{(t)}={\lvert \mathcal{D}_{k,y}^{(t)}\rvert}/{\lvert \mathcal{D}_{k}^{(t)}\rvert}$,
where $\mathcal{D}_{k,y}^{(t)}=\{x_i| c^{(t)}_i=k, y_i=y\}$.}

When most of samples in subgraph  $k$ belong to label $y$, {\em i.e.}, 
$r_{k,c}^{(t)}$ is over a threshold $\rho$, we add all the unlabeled samples in subgraph $k$ into labeled set $\mathcal{D}^L$ to form the updated $\mathcal{D}^{L^{(t)}}= \mathcal{D}^{L} \cup \mathcal{D}_{new}^{(t)}$,
where $\mathcal{D}_{new}^{(t)}=\{x_i | x_i \in \mathcal{D}_k^{(t)}, r_{k,y}^{(t)} > \rho\}$. 
Therefore, we can utilize $\mathcal{D}^{L^{(t)}}$ to update the label set $\mathcal{Y}^{(t)}$.

On the contrary, we denote the remaining unlabeled samples from $\mathcal{D}^U$ at epoch $t$ as $\mathcal{D}^{U^{(t)}}$. By the reconstructed dataset, we introduce a classification loss $\mathcal{L}_0$ and an equalization loss  $\mathcal{L}_1$ as follows:
\begin{align}
\mathcal{L}_0 &= -\sum\nolimits_t\sum\nolimits_{x_i\in \mathcal{D}^{L^{(t)}} }\sum\nolimits_{y_i\in \mathcal{Y}^{(t)}}
 y_i \cdot \log (p(\mathbf{y}|x_i)), \nonumber\\
\mathcal{L}_1 &= -\sum\nolimits_t\sum\nolimits_{x_i\in\mathcal{D}^{U^{(t)}}} v \cdot \log (p(\mathbf{y}|x_i)),
\end{align}
where $p(\mathbf{y}|x_i)$ denotes the prediction probability of $x_i$ over all ID classes,  $y_i$ is converted to a one-hot label vector of $x_i$, $v={\mathbf{1}_R}/{R}$ denotes the uniform posterior distribution over all of $R$ ID classes, where $\mathbf{1}_R$ is an all-ones vector with $R$-dimension.
Thus, $\mathcal{L}_1^{(t)}$  forces samples in $\mathcal{D}^{U^{(t)}}$ to uniformly distribute among $R$ ID classes. 

\subsection{Subgraph Aggregation for Cross-level Feature Update \label{sec:aggregation_function}}
{To update the cross-level image features, we design a subgraph aggregation strategy.
Especially, after label assignment, we can utilize these newly-labeled samples to update the feature encoder for better image features.
To obtain $G_{l+1}=\{V_{l+1}, E_{l+1}\}$, we first convert any subgraph $c^{(l)}_i$ in $G_{l}$ to node $v^{(l+1)}_i$ in $V_{l+1}$.
For $v^{(l+1)}_i$, we introduce its two  feature vectors: the identity feature $\tilde{f}^{(l+1)}_i$ and the average feature $\bar{f}^{(l+1)}_i$ defined as follows:}
\begin{equation}
    \tilde{f}^{(l+1)}_i  = \tilde{f}^{(l)}_{m_i} ~~~\mbox{and}~~~ \bar{f}^{(l+1)}_i  = \frac{1}{|c^{(l)}_i|}\sum\nolimits_{j\in c^{(l)}_i}   {\tilde{f}^{(l)}_{j}},
\label{eq:feature}
\end{equation}
where $m_i$ denotes the peak node index of the subgraph $c^{(l)}_i$, which is defined as: 
\begin{equation}
m_i = \operatorname*{argmax}_{j \in c^{(l)}_i} \hat{d}^{(l)}_{j}.
\end{equation}
Especially, for the first level, $\tilde{f}^{(0)}_i = \bar{f}^{(0)}_i = f_i$, where $f_i$ is the visual embedding feature. Thus, we utilize the identity feature $\tilde{f}^{(l)}_i$ to identify similar nodes across hierarchies, and use  the average feature $\bar{f}^{(l)}_i$ to extract the global information for all nodes in the subgraph.
The next-level input feature  of node $v^{(l+1)}_i$ is the concatenation of the peak feature and the average feature, {\em i.e.}, $ f^{l+1}_i = [\tilde{f}^{(l+1)}_i, \bar{f}^{(l+1)}_i]$.

\subsection{Data Augmentation for Model Generalization}
{In real-world applications, the supervised classification methods only aim to extract semantic representations to meet the minimum necessary, and they often overtrust the given  labels. However, in the challenging OOD detection task, we cannot overtrust  collected samples including OOD samples. We tend to widen the discrepancy between ID/OOD on similarity metric, which can strengthen the ability to guide ID/OOD samples to different subgraphs. By building an unsupervised training module, we can extract  enhanced features.}

Therefore, we leverage the data augmentation technology to obtain different augmented images. 
Then, for each image, we maximize the similarity between its different augmented versions. To obtain the augmented images of all the images, we first integrate $N^L$ labeled samples $\{x_i^L\}_{i=1}^{N^L}$ and $N^U$ unlabeled images $\{x_i^U\}_{j=1}^{N^U}$ into a total set $X=\{x_i\}_{i=1}^{N}$. Then, we augment any $x_i$ into two version:
$x_i^0$ and $x_i^1$, where $x_i^0=A^0(x_i)$ and $x_i^1=A^1(x_i)$; $A^0$ and $A^1$ denote two augmentations. Finally, we feed $x_i^0$ and $x_i^1$ into a shared feature encoder followed by an MLP layer. For convenience, we denote the augmented features as $b_i^0$ and $b_i^1$, respectively.

For the $t$-th epoch, we maximize the cosine similarity of the same samples in different augmented features ($\{b_i^0\}^{(t)}$ and $\{b_i^1\}^{(t)}$) to enhance the sample feature representation. Thus, we introduce the InfoNCE loss as the objective:
\begin{align}
\mathcal{L}_2=-\sum\nolimits_{i=1}^N \log (&\frac{\exp(cos(b_i^0, b_i^1))}{\sum_{j=1}^{N}\exp(cos(b_i^0, b_j))}\nonumber\\
&+\frac{\exp(cos(b_i^0, b_i^1))}{\sum_{g=1}^{N}\exp(cos(b_i^1, b_g))}),
\end{align}
where $b_j\in B^{1,(t)}$, $b_g\in B^{0,(t)}$, and $cos(\cdot,\cdot)$ denotes the cosine similarity function. 
Based on $\mathcal{L}_2$, we  map the enhanced representation into the lower-dimensional logit space, where  the class-specific similarity score can illustrate the semantic discrepancy of samples from different classes, and assign a more discriminative subgraph distribution to them.

Therefore,  our overall loss is as follows:
\begin{equation}\label{e:overall_loss}
\mathcal{L} = \mathcal{L}_0  + \beta \mathcal{L}_1 + \alpha \mathcal{L}_2 + \gamma \mathcal{L}_0^L,
\end{equation}
where $\alpha$, $\beta$ and $\gamma$ are parameters to balance different losses.

\subsection{Inference for Any Image}
Given an unlabeled image, a well-designed OOD detection model should correctly generate its prediction.
With a classification head $h_1$, we leverage a fully-connected layer to  map the image features into a logit space. We utilize the energy-based approach with temperature-scaled logits  for OOD detection. After training, we can utilize the similarity score from the logit $l(\mathbf{y}|x_i)=h_1(b_i)$ as the OOD score. 
 The similarity score is formulated as: 
\begin{align}\label{e:T-similarity}
g(x_i)=T \cdot \log\sum\nolimits_{c=1}^C \exp(\frac{l(y_c|x_i)}{T}),
\end{align}
where $l(y_m|x_i)$ is the logit of $x_i$ belonging to the ID class $y_m$, and ID samples have higher similarity score. 

Unlike traditional OOD detection methods \cite{liang2018enhancing,liu2020energy,sun2021react,lee2018simple,mohseni2020self,vyas2018out} that only train on ID samples, we pay more attention to OOD samples. These OOD samples will be more concentrated at minimum similarity due to the involvement of training OOD data under the SC-OOD setting. Thus, a proper temperature value $T$ can smooth the similarity distribution and  widen the discrepancy between ID/OOD on the similarity metric.

Therefore, we compare the similarity score to the predefined threshold $\delta$. Any image $x_i$ is classified as in-distribution if the softmax score  is greater than the threshold and vice versa. Mathematically, the out-of-distribution detector can be described as: 
\begin{align}
x_i \text{ is }\left\{\begin{matrix}\text{OOD sample,}&\text{ if } g(x_i)\le\delta,\\ 
\text{ID sample,}&\text{ if } g(x_i)>\delta, \end{matrix}\right.
\end{align}
where $\delta$ is chosen so that the true positive rate ({\em i.e.}, the fraction of ID images correctly classified as ID images) is 95\%.

\section{Experiments}\label{exp}

\subsection{Datasets}
Following \cite{yang2021semantically}, we use two SC-OOD benchmarks: CIFAR-10 and CIFAR-100. Each benchmark contains seven datasets: CIFAR-10 \cite{krizhevsky2009learning}, CIFAR-100 \cite{krizhevsky2009learning}, Texture \cite{cimpoi2014describing}, SVHN \cite{yuval2011reading}, Tiny-ImageNet \cite{le2015tiny}, LSUN \cite{yu2015lsun}  and Places365 \cite{zhou2017places}. 
Each benchmark treats a dataset (CIFAR-10 or CIFAR-100) as labeled dataset, and the others as unlabeled datasets.
The statistics of these datasets are as follows:
1)  CIFAR-10 benchmark: all the images in the CIFAR-10 dataset are ID samples; all the images in the CIFAR-100 dataset are OOD samples; all the images in the SVHN dataset are OOD samples; 1,207 images in the Tiny-ImageNet dataset are ID samples and 8,793 images are OOD samples; 2 images in the LSUN dataset are ID samples and 9,998 images are OOD samples; 1,305 images in the Places365 dataset are ID samples and 35,195 images are OOD samples.
2)  CIFAR-100 benchmark: all the images in the CIFAR-10 dataset are OOD samples; all the images in the CIFAR-100 dataset are ID samples; all the images in the SVHN dataset are OOD samples;  2,502 images in the Tiny-ImageNet dataset are ID samples and 7,498 images are OOD samples; 2,429 images in the LSUN dataset are ID samples and 7,571 images in the LSUN dataset are OOD samples; 2,727 images  in the Places365 dataset are ID samples and 33,773 images  are OOD samples.

\subsection{Evaluation Metrics}
For  fair comparison, we follow \cite{yang2021semantically} to evaluate the performance on both ID classification and OOD detection by the following  evaluation  metrics: 
1) FPR95: It computes the   False Positive Rate (FPR) value when the True Positive Rate (TPR) is 95\%, which denotes the rate of falsely recognized OOD when 95\% ID samples are recalled.
2) AUROC: Area Under the Receiver Operating Characteristic (AUROC) curve calculates the area by plotting the TPR as a function of the FPR. AUROC can serve as a threshold-independent metric to evaluate the OOD detection performance.
3) AUPR: Area Under the Precision-Recall (AUPR) curve measures  the area under the curve  by plotting recall  as a function of precision. Also, AUPR is a threshold-independent. Based on different  selections of positiveness, we consider two evaluation metrics: AUPR(In) (ID samples are positive) and AUPR(Out) (OOD samples are positive).
4) CCR@FPR$m$: It denotes Correct Classification Rate (CCR) for ID classification when the FPR equals $m$.
	CCR@FPR$m$ is used to evaluate OOD detection
 and ID classification  simultaneously.

\begin{table*}[th!]
\setlength{\tabcolsep}{11 pt}
\caption{\textbf{Performance comparison on the CIFAR-10 benchmark, where data partitions are identical for all methods.}
Our  AHGC achieves consistently the best performance on all OOD detection metrics.  $\uparrow$/$\downarrow$ means a higher/lower value is better.}
\label{T:CIFAR10_res18}
\centering
\small
\begin{tabular}{c|c|ccc|ccccc}
\toprule
\multirow{2}{*}{Dataset} & \multirow{2}{*}{Method}
& \multirow{2}{*}{FPR95~$\downarrow$}
& \multirow{2}{*}{AUROC~$\uparrow$}
& \multirow{2}{*}{AUPR(In/Out)~$\uparrow$}
& \multicolumn{4}{c}{CCR@FPR~$\uparrow$}
 \\
\cmidrule(lr){6-9}
& && && $10^{-4}$  & $10^{-3}$  & $10^{-2}$ & $10^{-1}$ \\
\midrule
\multirow{8}{*}{\begin{tabular}[c]{@{}c@{}}Texture \end{tabular}}
& MCD \cite{yu2019unsupervised}&83.92          & 81.59          & 90.20           / 63.27          & 4.97          & 10.51         & 29.52          & 62.10                   \\
&EBO  \cite{liu2020energy}   & 52.11          & 80.70          & 83.34          / 75.20          & 0.01          & 0.13          & 2.79          & 31.96                 \\
& OE \cite{hendrycksdeep} & 51.17          & 89.56          & 93.79          / 81.88          & 6.58           & 11.80           & 27.99          & 71.13                   \\
&ODIN  \cite{liang2018enhancing}   & 42.52          & 84.06          & 86.01          / 80.73          & 0.02          & 0.18          & 3.71          & 40.14                 \\
& {ConjNorm \cite{pengconjnorm}}& {23.78}& {95.59}& {97.40 / 95.88}& {18.42}& {45.29}& {70.96}& {89.73}\\
& UDG \cite{yang2021semantically}&  20.43          & 96.44          & 98.12          / 92.91          & 19.90           & 43.33          & 69.19          & 87.71                  \\
&Scone \cite{bai2023feed} & 18.26& 97.30& 97.88 / 93.11& 21.08& 40.92& 68.43& 88.29 \\
&\textbf{Our AHGC} & \textbf{0.89}&\textbf{99.42}&\textbf{99.67} / \textbf{99.05}&\textbf{54.03}&\textbf{75.02}&\textbf{87.81}&\textbf{94.62}         \\
\midrule
\multirow{8}{*}{\begin{tabular}[c]{@{}c@{}}SVHN \end{tabular}}
& MCD   \cite{yu2019unsupervised}      & 60.27          & 89.78          & 85.33         / 94.25          & 20.05         & 38.23         & 55.43          & 74.01                  \\
&ODIN  \cite{liang2018enhancing}    & 52.27          & 83.26          & 63.76          / 92.60          & 1.01          & 4.00          & 11.82         & 44.85                   \\
&EBO  \cite{liu2020energy}   & 30.56          & 92.08          & 80.95          / 96.28          & 1.85          & 5.74          & 21.44         & 75.81                   \\
& OE \cite{hendrycksdeep}&  20.88          & 96.43          & 93.62          / 98.32          & 32.72          & 47.33          & 67.20           & 86.75                  \\
& UDG \cite{yang2021semantically}& 13.26          & 97.49          & 95.66          / 98.69          & 36.64          & 56.81          & 76.77          & 89.54                 \\
&Scone \cite{bai2023feed}& 11.15& 95.48& 93.46 / 94.58& 32.18& 58.44& 72.38& 90.33 \\
&{ConjNorm  \cite{pengconjnorm}}& {9.89}& {96.32}& {94.88 / 96.25} & {33.82}& {67.94}& {80.31} & {91.72} \\
&\textbf{Our AHGC} &  \textbf{0.00}&\textbf{99.99}&\textbf{99.99} / \textbf{100.00}&\textbf{91.89}&\textbf{94.97}&\textbf{95.37}&\textbf{95.40}        \\
\midrule
\multirow{8}{*}{\begin{tabular}[c]{@{}c@{}}CIFAR-100 \end{tabular}}
& MCD     \cite{yu2019unsupervised}    & 74.00     & 82.78          & 83.97          / 79.16          & 0.80           & 4.99          & 18.88          & 58.18                    \\
& OE \cite{hendrycksdeep}& 58.54          & 86.22          & 86.17          / 84.88          & 3.64           & 6.55           & 19.04          & 61.11                   \\
&EBO  \cite{liu2020energy}   & 56.98          & 79.65          & 75.09          / 81.23          & 0.10          & 0.69          & 4.74          & 34.28                   \\
&ODIN  \cite{liang2018enhancing}    &  56.34          & 78.40          & 73.21          / 80.99          & 0.10          & 0.38          & 4.43          & 30.11                  \\
& UDG \cite{yang2021semantically}& 47.20           & 90.98          & 91.74          / 89.36          & 1.50            & 10.94          & 40.34          & 75.89                   \\
&Scone \cite{bai2023feed} & 43.79& 91.24 & 90.68 / 88.76 & 2.38 & 8.17& 42.96& 77.94 \\
&{ConjNorm \cite{pengconjnorm}}& {30.94}& {87.46}& {91.18 / 89.23}& {1.06}& {13.78}& {39.87}& {75.50}\\
& \textbf{Our AHGC} &  \textbf{24.82}&\textbf{92.72}&\textbf{93.50} / \textbf{91.69}&\textbf{3.81}&\textbf{21.00}&\textbf{47.87}&\textbf{80.04}      \\
\midrule
\multirow{8}{*}{\begin{tabular}[c]{@{}c@{}}Tiny-ImageNet \end{tabular}}
& MCD     \cite{yu2019unsupervised}    &   78.89          & 80.98          & 85.63          / 72.48          & 1.62          & 4.15          & 19.37          & 56.08                  \\
&ODIN \cite{liang2018enhancing}     & 59.09          & 79.69          & 79.34          / 77.52          & 0.36          & 0.63          & 4.49          & 34.52                   \\
& OE \cite{hendrycksdeep}& 58.98          & 87.65          & 90.90           / 82.16          & 14.37          & 18.84          & 33.65          & 66.03                  \\
&EBO   \cite{liu2020energy}  & 57.81          & 81.65          & 81.80          / 78.75          & 0.33          & 0.95          & 6.01          & 40.40                   \\
&Scone \cite{bai2023feed} & 55.62 & 93.48 & 95.34 / 89.59 & 10.76& 21.40 & 62.55& 81.64 \\
&{ConjNorm \cite{pengconjnorm}}& {52.80}& {90.74} & {83.25 / 80.24} & {9.52}& {19.37}& {59.88}& {79.45} \\
& UDG \cite{yang2021semantically}& 50.18          & 91.91          & 94.43          / 86.99          & 0.32           & 23.15          & 53.96          & 78.36                   \\
&\textbf{Our AHGC} &  \textbf{9.71}&\textbf{96.45}&\textbf{97.94} / \textbf{92.96}&\textbf{72.84}&\textbf{81.85}&\textbf{85.22}&\textbf{86.67}      \\
\midrule
\multirow{8}{*}{\begin{tabular}[c]{@{}c@{}}LSUN \end{tabular}}
& MCD   \cite{yu2019unsupervised}      & 68.96          & 84.71          & 85.74          / 81.50           & 1.75          & 7.93          & 21.88          & 61.54                  \\
& OE \cite{hendrycksdeep}& 57.97          & 86.75          & 87.69          / 85.07          & 11.8           & 19.62          & 29.22          & 61.95                  \\
&Scone \cite{bai2023feed}& 52.18& 94.33& 91.96 / 87.24& 12.90& 43.09& 68.99& 85.86 \\
&EBO \cite{liu2020energy}    &  50.56          & 85.04          & 82.80          / 85.29          & 0.24          & 1.96          & 11.35         & 50.43                   \\
& ODIN \cite{liang2018enhancing}     & 47.85          & 84.56          & 81.56          / 85.58          & 0.21          & 0.85          & 9.92          & 46.95                  \\
& UDG \cite{yang2021semantically}& 42.05          & 93.21          & 94.53          / 91.03          & 14.26          & 37.59          & 60.62          & 81.69                  \\
& {ConjNorm \cite{pengconjnorm}}& {38.29}& {93.67}& {95.82 / 92.89} & {13.84}& {41.95}& {64.31}& {88.72}\\
&\textbf{Our AHGC} &\textbf{1.72}&\textbf{98.71}&\textbf{98.84} / \textbf{98.63}&\textbf{53.97}&\textbf{69.41}&\textbf{82.55}&\textbf{93.14}        \\
\midrule
\multirow{8}{*}{\begin{tabular}[c]{@{}c@{}}Places365 \end{tabular}}
& MCD \cite{yu2019unsupervised}  & 72.08          & 83.51          & 69.44          / 92.52          & 3.29          & 7.97          & 23.07          & 60.22                  \\
& OE \cite{hendrycksdeep}& 55.64          & 87.00             & 73.11          / 94.67          & 11.36          & 17.36          & 26.33          & 62.23                  \\
& ODIN  \cite{liang2018enhancing}    & 53.94          & 82.01          & 54.92          / 93.30          & 0.47          & 1.68          & 7.13          & 39.63                   \\
&EBO  \cite{liu2020energy}   & 52.16          & 83.86          & 58.96          / 93.90          & 0.39          & 2.11          & 8.38          & 46.00                  \\
&Scone \cite{bai2023feed} & 48.63& 87.58& 80.49 / 92.93& 16.73& 42.76& 62.57& 81.53 \\
& UDG \cite{yang2021semantically}& 44.22          & 92.64          & 87.17          / 96.66          & 10.62          & 35.05          & 58.96          & 79.63                    \\
& {ConjNorm \cite{pengconjnorm}} & {36.02}& {93.58}& {86.72 / 95.85}& {12.40}& {40.10}& {60.70}& {82.34}  \\
&\textbf{Our AHGC} & \textbf{10.24}&\textbf{97.24}&\textbf{94.74} / \textbf{98.82}&\textbf{37.19}&\textbf{59.12}&\textbf{76.67}&\textbf{89.54}      \\
\midrule
\multirow{8}{*}{\begin{tabular}[c]{@{}c@{}}Mean \end{tabular}}
& MCD  \cite{yu2019unsupervised} & 73.02 & 83.89 & 83.39 / 80.53 & 5.41 & 12.30 & 28.02 & 62.02 \\
& ODIN  \cite{liang2018enhancing}    &   52.00 & 82.00 & 73.13 / 85.12 & 0.36 & 1.29 & 6.92 & 39.37 \\
& OE \cite{hendrycksdeep}& 50.53 & 88.93 & 87.55 / 87.83 & 13.41 & 20.25 & 33.91 & 68.20 \\
&EBO   \cite{liu2020energy}  &50.03 & 83.83 & 77.15 / 85.11 & 0.49 & 1.93 & 9.12 & 46.48  \\
&Scone \cite{bai2023feed}& 38.27& 93.24& 91.64 / 91.04& 16.01& 35.80& 62.98& 84.27 \\
& UDG \cite{yang2021semantically}& 36.22 & 93.78 & 93.61 / 92.61 & 13.87 & 34.48 & 59.97 & 82.14 \\
&{ConjNorm \cite{pengconjnorm}} & {31.95}& {92.89}& {91.54 / 91.72}& {14.84}& {38.07}& {62.67}& {84.58} \\
&\textbf{Our AHGC} & \textbf{7.90} &\textbf{97.42}&\textbf{97.45} / \textbf{96.86}&\textbf{52.29}&\textbf{66.90}&\textbf{79.25}&\textbf{89.90} \\
\bottomrule
\end{tabular}
\vspace{-12pt}
\end{table*}

\begin{table*}[th!]
\setlength{\tabcolsep}{11 pt}
\caption{\textbf{Performance comparison on the CIFAR-100 benchmark, where data partitions are identical for all methods.}
Our  AHGC achieves consistently the best performance on all OOD detection metrics.  $\uparrow$/$\downarrow$ means a higher/lower value is better.}
\vspace{-5pt}
\label{T:CIFAR100_res18}
\centering
\small
\begin{tabular}{c|c|ccc|ccccc}
\toprule
\multirow{2}{*}{Dataset} & \multirow{2}{*}{Method}
& \multirow{2}{*}{FPR95~$\downarrow$}
& \multirow{2}{*}{AUROC~$\uparrow$}
& \multirow{2}{*}{AUPR(In/Out)~$\uparrow$}
& \multicolumn{4}{c}{CCR@FPR~$\uparrow$} \\
\cmidrule(lr){6-9}
& && && $10^{-4}$  & $10^{-3}$  & $10^{-2}$ & $10^{-1}$ \\
\midrule
\multirow{8}{*}{\begin{tabular}[c]{@{}c@{}}Texture\end{tabular}}
& OE \cite{hendrycksdeep}& 86.56          & 73.89          & 84.48          / 54.84          & 0.66          & 2.86          & 12.86          & 41.81                 \\
& EBO \cite{liu2020energy}& 84.29          & 76.32          & 85.87          / 59.12          & 0.82          & 3.89          & 14.37          & 44.60                  \\
&MCD \cite{yu2019unsupervised}& 83.97          & 73.46          & 83.11          / 56.79          & 0.07          & 1.03          & 9.29           & 38.09                  \\
& ODIN \cite{liang2018enhancing} &  79.47          & 77.92          & 86.69          / 62.97          & \textbf{2.66}          & 4.66          & 15.09          & 45.82                   \\
&Scone \cite{bai2023feed}& 78.34& 81.25& 86.98 / 61.36& 0.75& 4.98& 16.74& 48.76 \\
&{ConjNorm \cite{pengconjnorm}}& {77.21}& {76.39}& {85.72 / 62.83}& {0.45}& {3.27}& {11.80}& {42.74}\\
& UDG \cite{yang2021semantically}&  75.04          & 79.53          & 87.63          / 65.49          & 1.97          & 4.36          & 9.49           & 33.84                \\
&\textbf{Our AHGC} &  \textbf{22.93}& \textbf{84.64}& \textbf{91.58} / \textbf{67.42}&0.91&\textbf{9.41}&\textbf{30.64}&\textbf{56.73}      \\
\midrule
\multirow{8}{*}{\begin{tabular}[c]{@{}c@{}}SVHN\end{tabular}}
& ODIN \cite{liang2018enhancing} & 90.33          & 75.59          & 65.25          / 84.49          & 4.98          & 12.02         & 23.79          & 46.61                 \\
& MCD \cite{yu2019unsupervised}& 85.82          & 76.61          & 65.50          / 85.52          & 3.03          & 8.66          & 23.15          & 45.44                  \\
& EBO \cite{liu2020energy}& 78.23          & 83.57          & 75.61          / 90.24          & 9.67          & 17.27         & 33.70          & 57.26                   \\
& OE \cite{hendrycksdeep}& 68.87          & 84.23          & 75.11          / 91.41          & 7.33          & 14.07         & 31.53          & 54.62                   \\
& UDG \cite{yang2021semantically}& 60.00          & 88.25          & 81.46          / 93.63          & 14.90         & 25.50         & 38.79          & 56.46                 \\
&{ConjNorm \cite{pengconjnorm}}& {59.32}& {82.17}& {79.32 / 90.44}& {13.05}& {23.85}& {30.92}& {58.40} \\
&Scone \cite{bai2023feed} & 58.73& 90.62& 83.97 / 92.39& 12.92 & 20.76& 28.63& 60.79 \\
&\textbf{Our AHGC} & \textbf{0.46}          & \textbf{97.70}          & \textbf{97.75}  / \textbf{97.36}         & \textbf{61.34}        & \textbf{69.62}         & \textbf{72.82}          & \textbf{74.36}                   \\
\midrule
\multirow{8}{*}{\begin{tabular}[c]{@{}c@{}}CIFAR-10\end{tabular}}
& MCD \cite{yu2019unsupervised}& 87.74          & 73.15          & 76.51          / 67.24          & 0.35          & 3.26          & 16.18          & 41.41                   \\
&{ConjNorm \cite{pengconjnorm}}& {85.87}& {72.55}& {75.50 / 65.38}& {1.29}& {4.76}& {13.85}& {42.17} \\
& UDG \cite{yang2021semantically}& 83.35          & 76.18          & 78.92          / 71.15          & 1.99          & 5.58          & 17.27          & 42.11                \\
& ODIN \cite{liang2018enhancing} & 81.82          & 77.90          & 79.93          / 73.39          & 0.09          & 3.69          & 15.39          & 47.20                   \\
&Scone \cite{bai2023feed} & 81.38& 72.34& 73.11 / 64.88& 1.09& 4.30& 15.33& 42.95\\
& EBO \cite{liu2020energy}& 81.25          & \textbf{78.95}          & 80.01          / \textbf{74.44}          & 0.05          & 4.63          & 18.03          & 48.67 \\
& OE \cite{hendrycksdeep}& 79.72          & 78.92          & \textbf{81.95}          / 74.28          & \textbf{2.82}          & \textbf{9.53}      & \textbf{23.90}   & \textbf{48.21}   \\
&\textbf{Our AHGC} & \textbf{43.98}          & 74.47          & 78.21    /  67.13          & 2.46          & 6.88         & 16.96          & 44.42                 \\
\midrule
\multirow{8}{*}{\begin{tabular}[c]{@{}c@{}}Tiny-ImageNet\end{tabular}}
& MCD \cite{yu2019unsupervised}& 84.46          & 75.32          & 85.11          / 59.49          & 0.24          & 6.14          & 19.66          & 41.44                   \\
&OE \cite{hendrycksdeep}& 83.41          & 76.99          & 86.36          / 60.56          & 0.22          & 8.50          & 21.95          & 43.98                  \\
& EBO \cite{liu2020energy}& 83.32          & 78.34          & 87.08          / 62.13          & 1.04          & 6.37          & 21.44          & 47.92                  \\
&Scone \cite{bai2023feed} & 82.92& 80.87& 88.93 / 63.90& 2.36& 10.62& 23.46& 49.38 \\
& ODIN \cite{liang2018enhancing} & 82.74          & 77.58          & 86.26          / 61.38          & 0.20          & 3.78          & 15.99          & 45.56                   \\
&{ConjNorm \cite{pengconjnorm}}& {82.30}& {79.42}& {87.96 / 60.44}& {1.73}& {13.59}& {20.88}& {44.10}\\
& UDG \cite{yang2021semantically}& 81.73          & 77.18          & 86.00          / 61.67          & 0.67          & 4.82          & 17.80          & 41.72                   \\
&\textbf{Our AHGC} & \textbf{0.49}          & \textbf{89.19}          & \textbf{95.01}          / \textbf{73.42}          & \textbf{50.02}          & \textbf{57.76}          & \textbf{59.66}          & \textbf{60.79}              \\
\midrule
\multirow{8}{*}{\begin{tabular}[c]{@{}c@{}}LSUN\end{tabular}}
& MCD \cite{yu2019unsupervised}& 86.08          & 74.05          & 84.21          / 58.62          & 1.57          & 5.16          & 18.05          & 41.25                  \\
& EBO \cite{liu2020energy}& 84.51          & 77.66          & 86.42          / 61.40          & 1.59          & 6.44          & 19.58          & 46.66                  \\
&OE \cite{hendrycksdeep}& 83.53          & 77.10          & 86.28          / 60.97          & 1.72          & 7.91          & 22.61          & 44.19                   \\
& ODIN \cite{liang2018enhancing} & 80.57          & 78.22          & 86.34          / 63.44          & 1.68          & 5.59          & 17.37          & 45.56                   \\
&{ConjNorm \cite{pengconjnorm}}& {79.82}& {77.29}& {85.17 / 61.98}& {2.84}& {6.08}& {20.95}& {44.88}\\
& UDG \cite{yang2021semantically}&  78.70          & 76.79          & 84.74          / 63.05          & 1.59          & 5.34          & 18.04          & 44.70                \\
&Scone \cite{bai2023feed} & 75.09& 78.65& 85.06 / 60.29& 3.89& 7.63 & 25.95& 45.63  \\
&\textbf{Our AHGC} & \textbf{25.15}         & \textbf{80.42}          & \textbf{88.88}          / \textbf{62.18}          & \textbf{7.56}         & \textbf{19.15}          & \textbf{30.26}         & \textbf{49.71}                    \\
\midrule
\multirow{8}{*}{\begin{tabular}[c]{@{}c@{}}Places365\end{tabular}}
& MCD \cite{yu2019unsupervised}& 82.74          & 76.30          & 61.15          / 87.19          & 1.08          & 3.35          & 14.04          & 43.37                 \\
& EBO \cite{liu2020energy}& 78.37          & 80.99          & 68.22          / 89.60          & 1.40          & 4.94          & 21.32          & 51.21                   \\
&OE \cite{hendrycksdeep}& 78.24          & 79.62          & 67.13          / 88.89          & 3.69          & 7.35          & 20.22          & 47.68                   \\
& ODIN \cite{liang2018enhancing} & 76.42          & 80.66          & 66.77          / 89.66          & 1.45          & 4.16          & 18.98          & 49.60                  \\
& UDG \cite{yang2021semantically}& 73.86          & 79.87          & 65.36          / 89.60          & 1.96          & 6.33          & 22.03          & 47.97                   \\
&{ConjNorm \cite{pengconjnorm}}& {73.34}& {78.86}& {67.40 / 88.27}& {2.87}& {7.34}& {21.99}& {50.43}\\
&Scone \cite{bai2023feed} & 72.44& 80.96& 68.65 / 88.32& 2.96& 8.06& 24.95& 51.64 \\
&\textbf{Our AHGC} & \textbf{27.10}          & \textbf{81.73}          & \textbf{71.82} / \textbf{89.79}         & \textbf{3.93}          & \textbf{10.93}          & \textbf{27.15}          & \textbf{52.44}                 \\
\midrule
\multirow{8}{*}{\begin{tabular}[c]{@{}c@{}}Mean\end{tabular}}
& MCD \cite{yu2019unsupervised}& 85.14 & 74.82 & 75.93 / 69.14 & 1.06 & 4.60 & 16.73 & 41.83  \\
& ODIN \cite{liang2018enhancing} & 81.89 & 77.98 & 78.54 / 72.56 & 1.84 & 5.65 & 17.77 & 46.73 \\
& EBO \cite{liu2020energy}& 81.66 & 79.31 & 80.54 / 72.82 & 2.43 & 7.26 & 21.41 & 49.39  \\
&OE \cite{hendrycksdeep}& 80.06 & 78.46 & 80.22 / 71.83 & 2.74 & 8.37 & 22.18 & 46.75  \\
&Scone \cite{bai2023feed} & 76.36& 78.06& 79.96 / 70.50& 3.46& 8.75& 18.99& 49.33 \\
&{ConjNorm \cite{pengconjnorm}}& {76.31}& {77.78}& {80.18 / 71.56}& {3.71}& {9.82}& {20.07}& {47.12} \\
& UDG \cite{yang2021semantically}& 75.45 & 79.63 & 80.69 / 74.10 & 3.85 & 8.66 & 20.57 & 44.47  \\
&\textbf{Our AHGC} & \textbf{20.02} & \textbf{84.69}  & \textbf{87.21} / \textbf{76.22}  &  \textbf{21.04}  & \textbf{28.96} &  \textbf{39.58} &  \textbf{56.41}   \\
\bottomrule
\end{tabular}
\vspace{-12pt}
\end{table*}

\subsection{Experimental Settings}
In this paper, we utilize the standard ResNet-18 network as the feature encoder. We use the standard SGD optimizer  with the weight decay of $0.0005$ and the momentum of $0.9$ to train it on two SC-OOD benchmarks: the CIFAR-10 benchmark and the CIFAR-100 benchmark. 
The two benchmarks treat CIFAR-10 and CIFAR-100 as the  labeled dataset in their corresponding benchmarks, respectively. Besides, the Tiny-ImageNet training dataset is utilized as the  external unlabeled dataset $\mathcal{D}^U$ in both benchmarks during the training process. We utilize a cosine learning rate scheduler with an initial learning rate of 0.1, taking totally 180 epochs. For the labeled dataset $\mathcal{D}^L$, its batch size is 64, while the batch size for the unlabeled dataset $\mathcal{D}^U$ is 128. The total subgraph number is $1500$ and $1800$ for  CIFAR-10 and CIFAR-100 benchmarks, respectively. For hyper-parameters, we set $\alpha=0.15,\beta=0.5,\gamma=10^{-4}$. All  experiments are implemented by PyTorch.

\subsection{Comparison with State-of-the-arts}

To evaluate the effectiveness of our proposed AHGC, we follow \cite{yang2021semantically} to compare  AHGC with the following open-source state-of-the-art OOD detection methods:
The following state-of-the-art OOD detection methods are compared:
1) ODIN \cite{liang2018enhancing} employs the temperature scaling and input perturbation in the classification framework. 2) EBO \cite{liu2020energy} proposes an energy-based framework, where the softmax confidence score is replaced by the energy score. 3) MCD \cite{yu2019unsupervised} first utilizes a two-head CNN network as the classifier and then maximizes the discrepancy between the two classifiers for OOD detection. 4) OE \cite{hendrycksdeep} leverages OOD samples to teach a network heuristics for OOD detection. 5) UDG \cite{yang2021semantically} introduces a concise framework based on K-means clustering with the help of  an external unlabeled set. 
6) Scone \cite{bai2023feed} designs a margin-based learning framework that leverages freely available unlabeled data in the wild to capture the environmental test-time OOD distributions under both covariate and semantic shifts. 
{7) ConjNorm \cite{pengconjnorm} introduce a novel Bregman divergence-based theoretical framework to design density functions in exponential family distributions.
ODIN, EBO, MCD,  Scone and ConjNorm directly treat one labeled dataset as ID and all unlabeled datasets as OOD. UDG and our AHGC treat some unlabeled samples shared similar semantics with labeled samples as ID.}

For convenience, we directly cite some results in \cite{yang2021semantically} from corresponding works, which are trained and tested on the challenging benchmarks. As shown in Table \ref{T:CIFAR10_res18} and Table \ref{T:CIFAR100_res18}, we compare our proposed AHGC with state-of-the-art methods, where AHGC reaches the best results in  most cases. Particularly, when we treat the Tiny-ImageNet as the unlabeled dataset, our  AHGC achieves the following amazing improvement compared with the second best method UDG: 1) In the CIAFR-10 benchmark,  AHGC  improves the performance by 40.47\% in terms of ``FPR95''.  2) In the CIAFR-100 benchmark,  AHGC outperforms compared methods by 81.24\% in terms of ``FPR95''.
The main reason is that our AHGC can well understand the latent semantics between multi-granularity images on different datasets for OOD detection, while UDG only conducts K-means clustering on different images, which  fails to understand the latent  relationship between labeled
and unlabeled samples with different granularity.

{\noindent\textbf{Performance on the CIFAR-10 benchmark.} We compare our proposed AHGC with the state-of-the-art  methods on the CIFAR-10 benchmark in Table \ref{T:CIFAR10_res18}, where our  AHGC outperforms all compared methods with a large margin over all the metrics. For the mean of six unlabeled datasets, compared with the best compared method ConjNorm, our AHGC  achieves performance improvements 24.05\%, 4.53\%, 5.91\%, 5.14\%, 37.45\%, 28.83\%, 16.58\% and 5.32\% on  all the metrics, respectively. 
For the Texture dataset, our AHGC outperforms Scone  by 34.10\% over ``CCR@FPR$10^{-3}$''. For the SVHN dataset, our AHGC outperforms ConjNorm by 58.07\% over ``CCR@FPR$10^{-4}$''. For the CIFAR-100 dataset, our AHGC outperforms ConjNorm by 8.00\% over ``CCR@FPR$10^{-3}$''.
For the Tiny-ImageNet dataset, our AHGC outperforms UDG  by 72.52\% over ``CCR@FPR$10^{-4}$''. 
For the LSUN dataset, our AHGC outperforms ConjNorm by 40.13\% over ``CCR@FPR$10^{-4}$''. 
 As for the Places365 dataset, our AHGC beats ConjNorm by 25.78\% over ``FPR95''. The significant improvement illustrates the effectiveness of
our AHGC.}

{\noindent\textbf{Performance on the CIFAR-100 benchmark.} In Table~\ref{T:CIFAR100_res18},
we also compare our proposed AHGC with these state-of-the-art  methods on the CIFAR-100 benchmark, where our  AHGC outperforms all compared methods by a large margin in most cases. For the mean of six unlabeled datasets, compared with the best compared method UDG, our AHGC  achieves 55.43\%, 5.06\%, 6.52\%, 1.95\%, 17.19\%, 20.30\%, 19.01\% and 11.94\% improvements on  all the metrics, respectively. For the Texture dataset, our AHGC outperforms UDG  by 52.11\% over ``FPR95''. For the SVHN dataset, our AHGC beats Scone  by 58.27\% over ``FPR95''. For the CIFAR-10 dataset, our AHGC outperforms OE  by 35.74\% over ``FPR95''. For the Tiny-ImageNet dataset, our AHGC outperforms UDG  by 81.24\% over ``FPR95''. For the LSUN dataset, our AHGC outperforms Scone  by 49.94\% over ``FPR95''. As for the Places365 dataset, our AHGC outperforms Scone  by 45.34\% over ``FPR95''. The significant improvement shows the effectiveness of our  AHGC.}

\begin{figure*}[t!]
    \centering
   \includegraphics[width=0.25\textwidth]{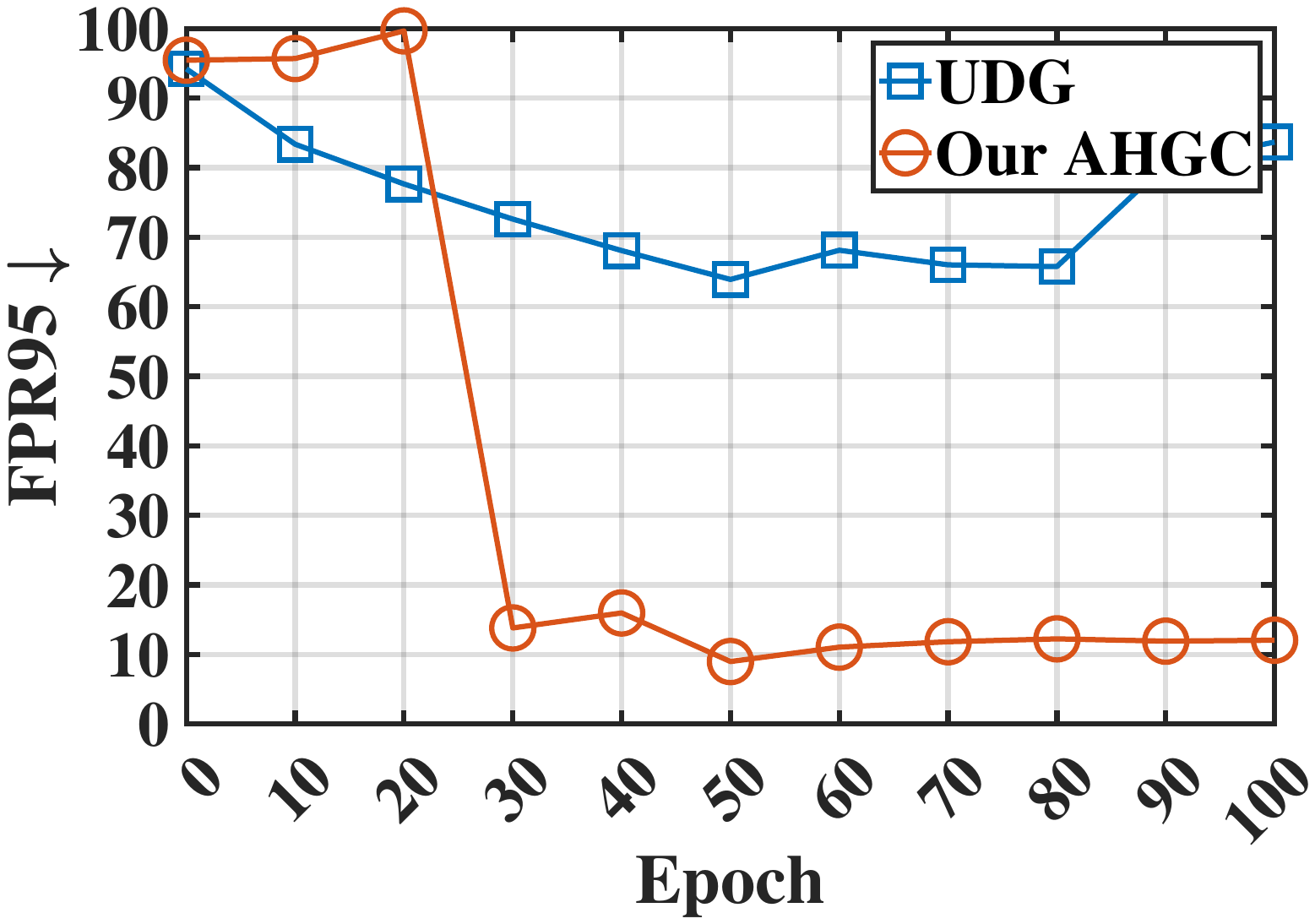}
\hspace{-0.10in}
\includegraphics[width=0.25\textwidth]{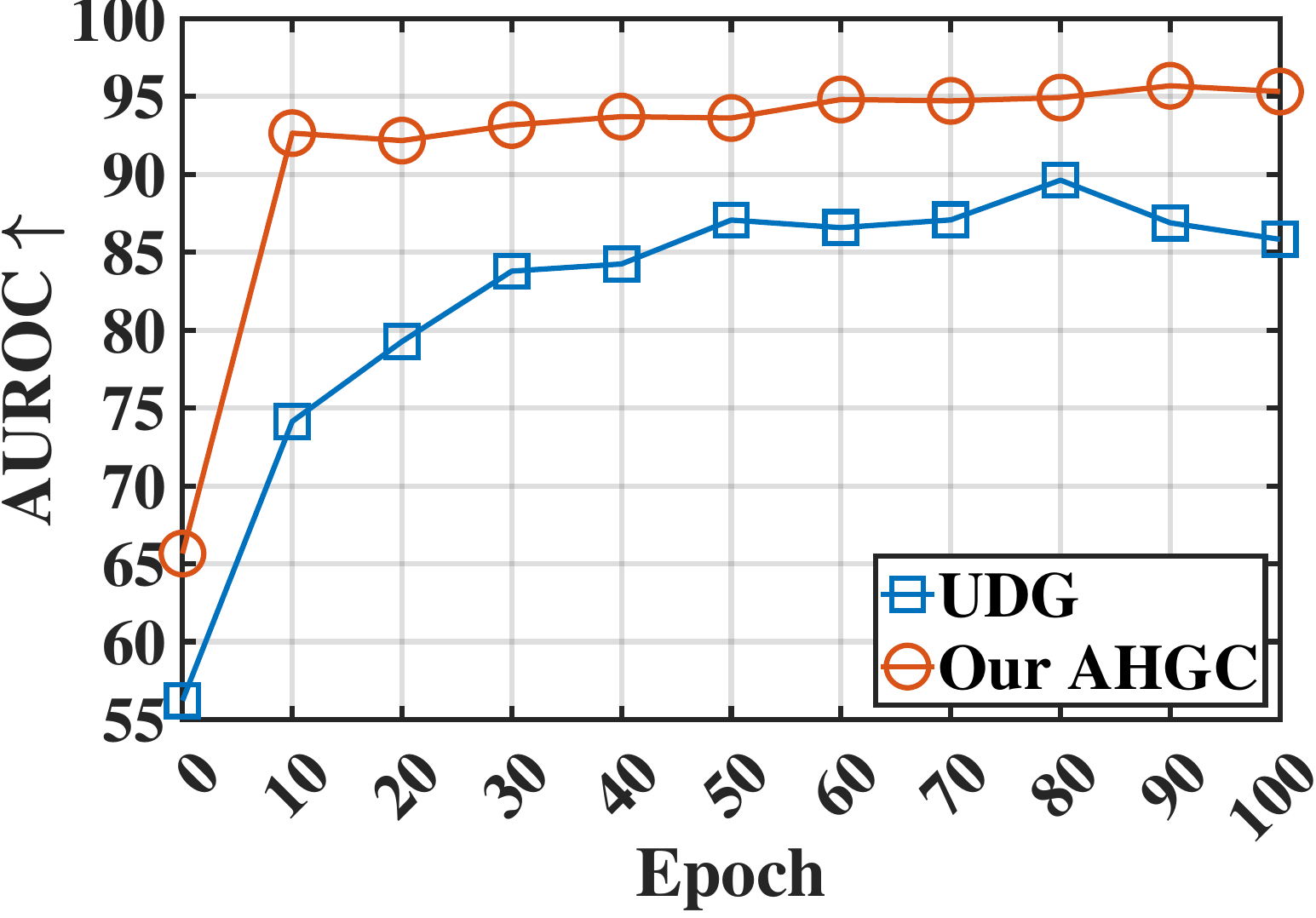}
\hspace{-0.10in}
\includegraphics[width=0.25\textwidth]{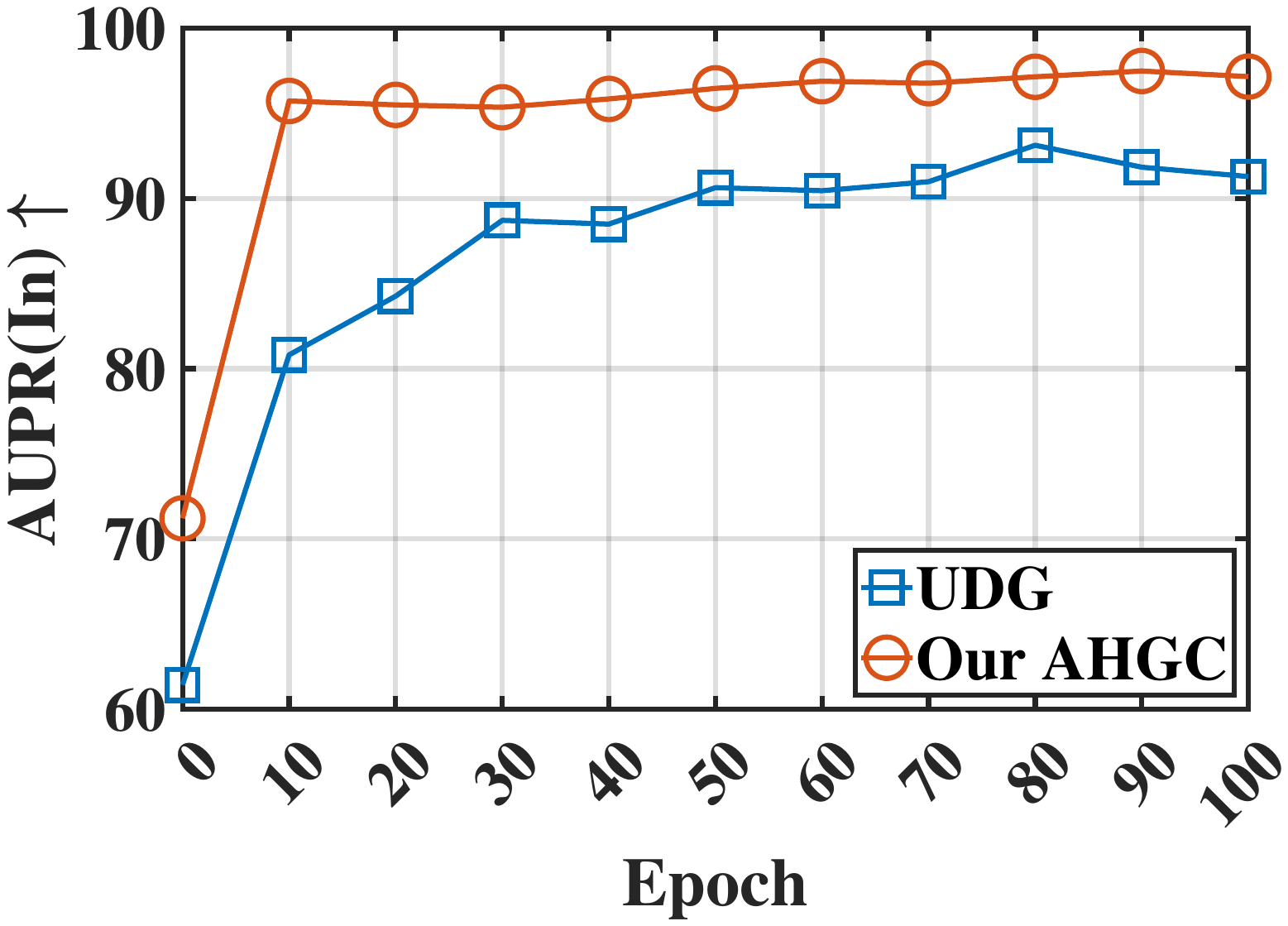}
\hspace{-0.10in}
\includegraphics[width=0.25\textwidth]{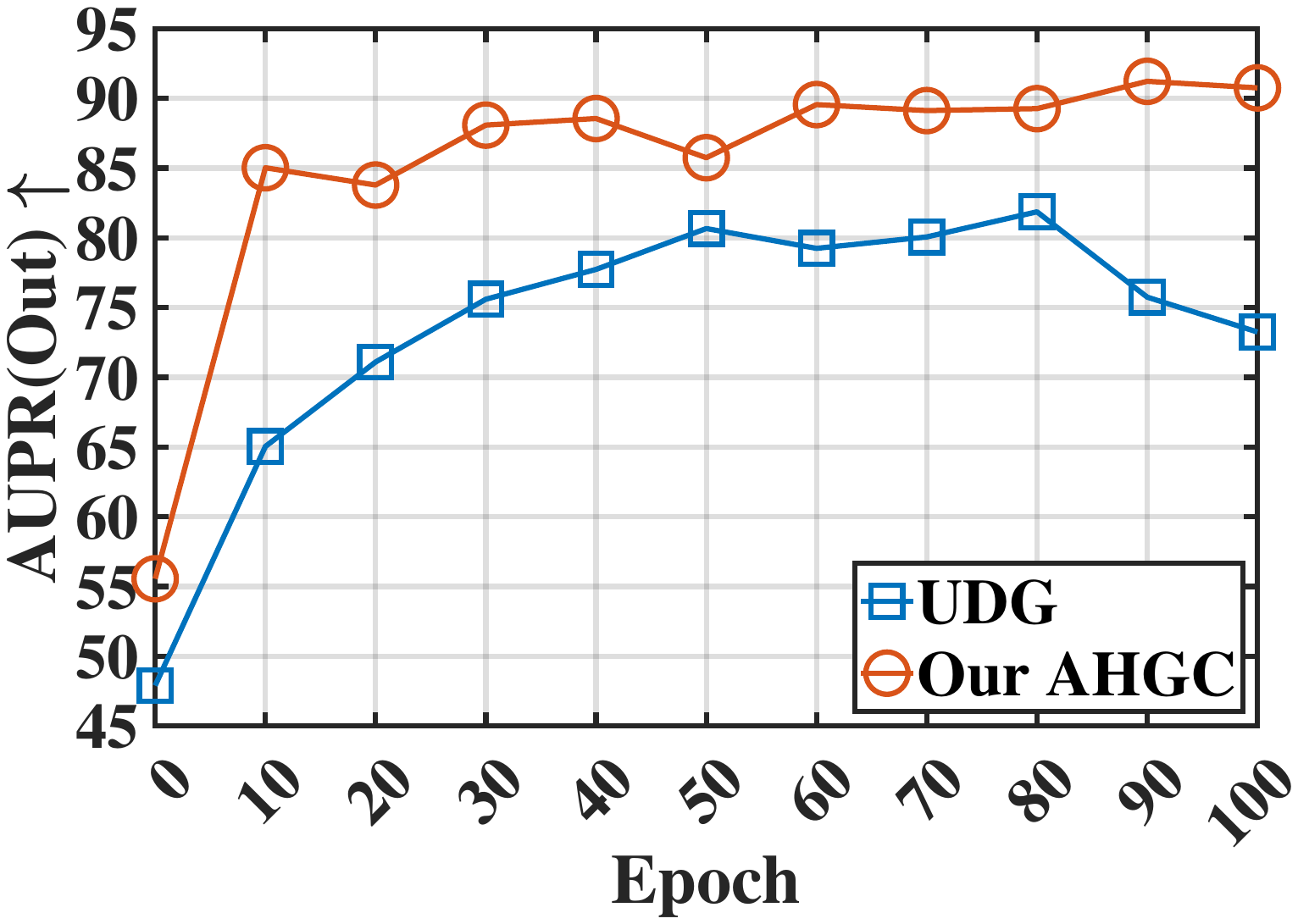}
	\caption{Performance of our proposed AHGC and UDG \cite{yang2021semantically} on  CIFAR-10 benchmark  during training process. Since the maximum number of epochs on UDG is 100, we report our performance of the first 100 epochs for fair comparison.}
	\label{fig:train_process}
 \vspace{-5pt}
\end{figure*}

\noindent \textbf{Analysis.} From the above results, we can observe that: 
(i) In all the datasets, our proposed AHGC on these two benchmarks (CIFAR-10 and CIFAR-100) outperforms  these state-of-the-art methods by a large margin in most cases. There are two main reasons as follows: Firstly, our  AHGC  is a more effective model, which leverages the graph cut to explore the latent  relationship of different samples under the same semantics. Secondly, we focus on the challenging SC-OOD setting, while many compared methods directly treat one labeled dataset as ID and all unlabeled datasets as OOD. The better performance in the  SC-OOD setting shows the effectiveness of our AHGC.
(ii) Although UDG is also under the SC-OOD setting, it is defeated by Scone in many cases because UDG cannot mine the latent relationship between multi-granularity images by simple K-means clustering. 
(iii) In Table~\ref{T:CIFAR100_res18}, for the CIFAR-10 dataset, two traditional methods (EBO and OE) perform well in terms of some metrics. The main reason is that all the images in  the CIFAR-100 dataset disjoint from the images in the CIFAR-10 dataset, which exactly meets their assumptions that labeled CIFAR-100 images are ID and unlabeled CIFAR-10 images are OOD. However, EBO and OE achieve unsatisfactory performance in the other unlabeled datasets (Texture, SVHN, Tiny-ImageNet, LSUN, Places365). This illustrates the significance of mining the unlabeled ID samples.
(iv) These  compared methods generally perform well on SVHN and Texture datasets, while they often suffer a defeat on the Tiny-ImageNet, LSUN and Places365 datasets.
It is because the images on SVHN and Texture datasets
have relatively flat backgrounds, which are quite different in style from those in CIFAR10/CIFAR-100, and the resulting covariate shifts make it easier for the model to identify them as OOD examples. Our AHGC especially achieves performance improvement on all the datasets in most cases.
This indicates that more
exploration between multi-granularity images is required to overcome the interference of covariate shifts in the OOD detection tasks.
{Besides, since our proposed AHGC satisfies the learnability condition \cite{fang2022out,fang2024learnability}, AHGC can effectively address the  multi-granularity OOD detection task. }

\noindent \textbf{Performance comparison during training process.} To further analyze the performance of UDG and our proposed AHGC, we report their performance during the training process. For convenience, we report their performance in every  10 epochs.
Fig. \ref{fig:train_process} shows the results. 
Obviously, our AHGC outperforms UDG with  large margins after 10 epochs for all the cases, which shows the effectiveness of our AHGC. Besides, our AHGC can converge faster than UDG.

\begin{table}[t!]
    \caption{Main ablation study on CIFAR-10 and CIFAR-100 benchmarks, where we utilize energy-based similarity score with temperature scaling to distinguish ID/OOD samples. We report the mean values of all unlabeled datasets.}
    \vspace{-5pt}
	\centering
 \small
	  \setlength{\tabcolsep}{9.5 pt}
	\begin{tabular}{c| c c c c c c c c c}
		\toprule
  \multicolumn{4}{c}{CIFAR-10 as the labeled dataset}\\
\midrule
         Modules& FPR95~$\downarrow$ & AUROC~$\uparrow$ & AUPR(In/Out)~$\uparrow$\\
        \hline
        AHGC(a)& 41.28& 83.80& 83.46 / 80.12\\
        AHGC(b)& 18.45& 87.36& 86.47 / 88.72\\
        AHGC(c)&13.78& 91.30& 92.57 / 92.34\\
        \textbf{AHGC(full)}&\textbf{7.90} &\textbf{97.42}&\textbf{97.45} / \textbf{96.86}\\
        \bottomrule
		\toprule
  \multicolumn{4}{c}{CIFAR-100 as the labeled dataset}\\
\midrule
         Modules& FPR95~$\downarrow$ & AUROC~$\uparrow$ & AUPR(In/Out)~$\uparrow$\\
        \hline
        AHGC(a)& 57.32& 73.51& 72.30 / 60.28\\
        AHGC(b)& 42.73& 79.72& 78.32/ 65.75\\
        AHGC(c)& 37.48& 80.36& 81.35 / 68.42\\
        \textbf{AHGC(full)}&\textbf{20.02} & \textbf{84.69}  & \textbf{87.21} / \textbf{76.22}\\
        \bottomrule
	\end{tabular}
	\label{tab:main_ablation}
 \vspace{-5pt}
\end{table}

\subsection{Ablation Study}

\subsubsection{Main ablation study} 
As shown in Table~\ref{tab:main_ablation}, we first conduct the main ablation study to examine the effectiveness of all the main modules in our model. Therefore, we design the following ablation models: (i) AHGC(a): we only utilize the classification loss $\mathcal{L}_0$ for OOD detection. (ii) 
AHGC(b): we combine  $\mathcal{L}_0$ and $\mathcal{L}_1$ for OOD detection.
(iii) AHGC(c):  we joint the classification loss $\mathcal{L}_0$ and the InfoNCE loss $\mathcal{L}_2$ in data augmentation for OOD detection. (iv) AHGC(full): we use our full AHGC model.

Obviously, our AHGC(full) outperforms the other ablation models on both benchmarks, which shows the effectiveness of each module. On the CIFAR-10 benchmark, compared with AHGC(a), AHGC(full) brings 33.38\% improvement over ``FPR95''; compared with AHGC(b), AHGC(full) raises the performance around 10.55\% over ``FPR95''; compared with AHGC(c), AHGC(full) obtains the performance improvement about 5.88\% over ``FPR95''.  On the CIFAR-100 benchmark, compared with AHGC(a), AHGC(full) brings 37.30\% improvement over `FPR95''; compared with AHGC(b), AHGC(full) raises the performance about 22.71\% over ``FPR95''; compared with AHGC(c), AHGC(full) obtains the performance improvement around 17.46\% over ``FPR95''.

It is because our subgraph aggregation  module can close the  discrepancy between labeled  and unlabeled images with the same semantics to transfer the label knowledge; our intra-subgraph label assignment module can reduce the semantic gaps between different datasets in each subgraph semantics; the data augmentation module is able to learn more fine-grained and representative image features for graph cut different datasets.

Based on the above results, we can obtain the following conclusions: 1) Compared with each model, the full model performs the best on all the datasets, which illustrates the effectiveness of each alignment module of our AHGC. It shows that our AHGC can effectively handle the OOD detection task. Meanwhile, it also illustrates that our full model is robust to address this OOD detection task. 2) We can observe that both AHGC(b) and AHGC(c) outperforms Baseline. It is because  each loss is important for the OOD detection task to  transfer the labels from labeled datasets to unlabeled datasets.

\begin{figure}[t!]
    \centering
\includegraphics[width=0.24\textwidth]{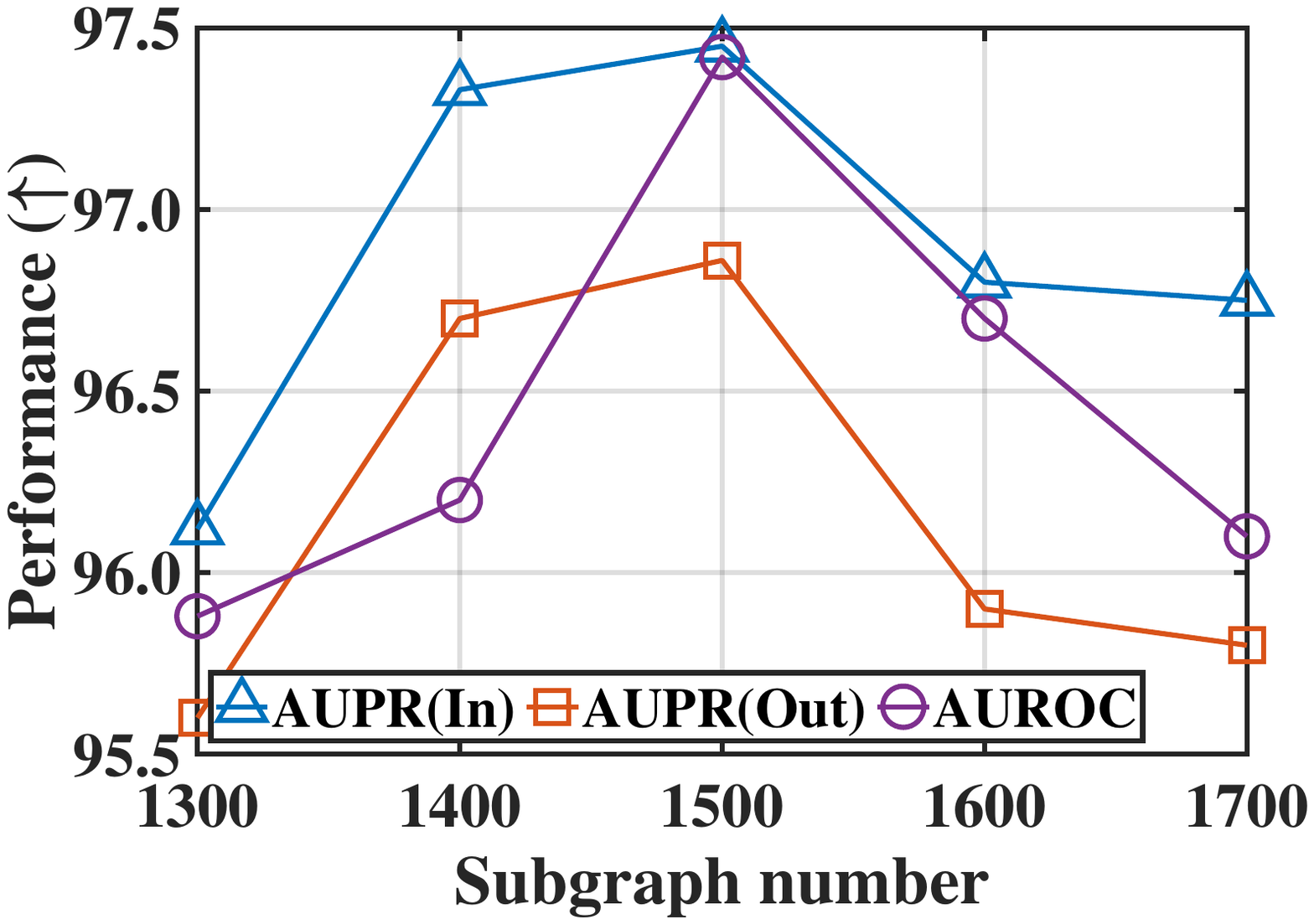}
\hspace{-0.10in}
\includegraphics[width=0.24\textwidth]{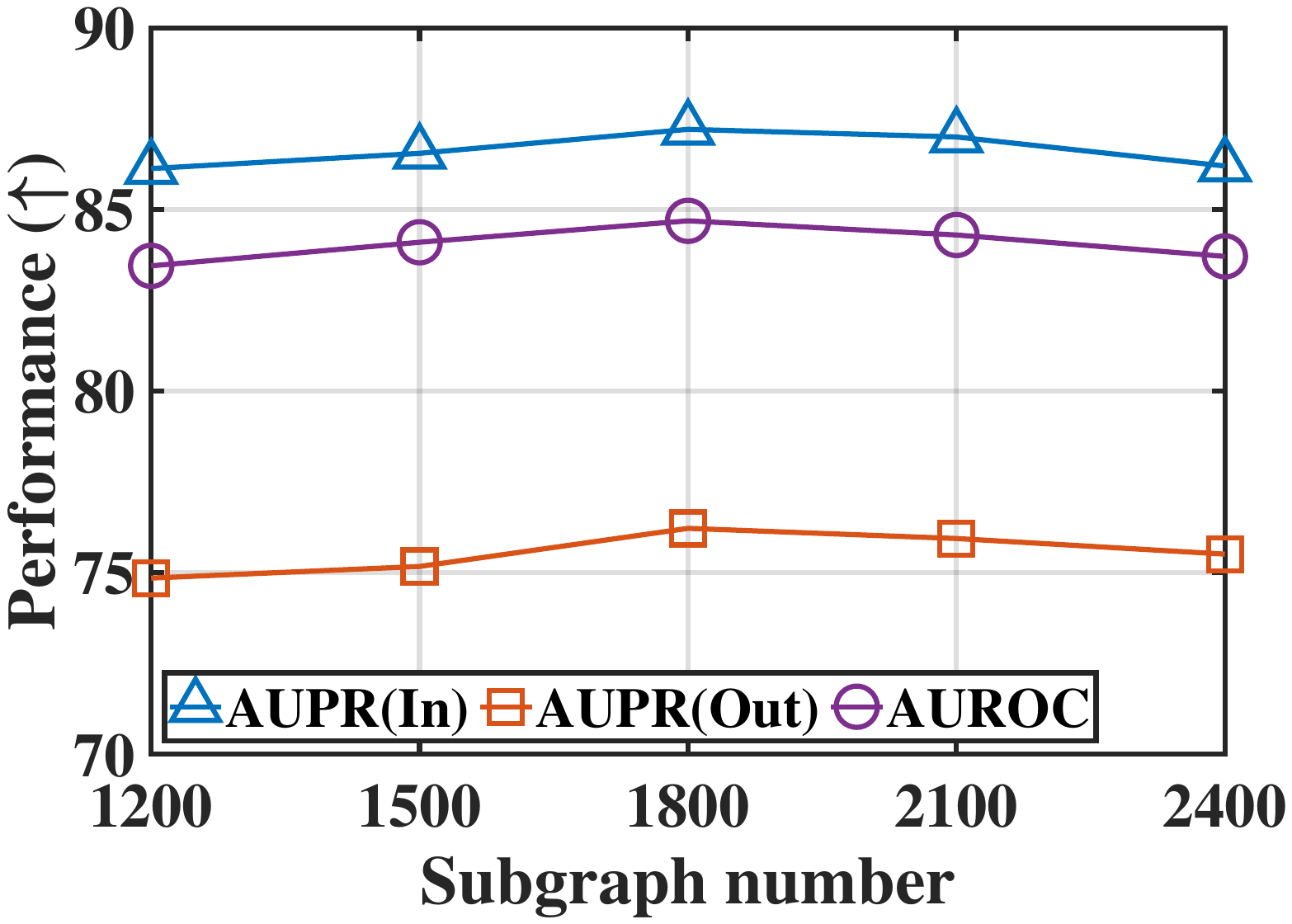}
	\caption{Effect of subgraph number $K$ on the CIFAR-10 benchmark (left) and for the CIFAR-100 benchmark (right).}
	\label{fig:subgraph_no}
 \vspace{-5pt}
\end{figure}

\subsubsection{Effect of subgraph number $K$} 
We further conduct  ablation study to analyze the impact of the subgraph numbers $K$. As shown in Fig. \ref{fig:subgraph_no}, we could observe that, with the increase of $K$, the variation of the performance follows a general trend, {\em i.e.}, rises at first and then starts to decline. It is because too small $K$ will lead to the case that many unlabeled OOD samples are clustered into a subgraph with a high rate of ID samples, which makes many OOD samples  mistakenly assigned improper labels. When $K$ is very large, many unlabeled ID samples are grouped into a subgraph, which consists of a few ID samples. It will result in the case that these unlabeled ID samples cannot obtain labels and be divided into OOD samples. Both cases will lead to unsatisfactory performance.
The optimal subgraphs number is $K=1500$ for the CIFAR-10 benchmark and $K=1800$ for the CIFAR-100 benchmark.

\begin{table*}[t!]
\caption{Ablation study on feature encoders on the  CIFAR-10 benchmark, where ``ACC'' means classification accuracy. 
 }
 \vspace{-5pt}
\label{T:diff_feature_encoder}
\centering
\small
\setlength{\tabcolsep}{8 pt}
\begin{tabular}{c|c|ccc|cccc|cccccccc}
\toprule
\multirow{2}{*}{Dataset} & \multirow{2}{*}{Feature encoder}
& \multirow{2}{*}{FPR95~$\downarrow$}
& \multirow{2}{*}{AUROC~$\uparrow$}
& \multirow{2}{*}{AUPR(In/Out)~$\uparrow$}
& \multicolumn{4}{c|}{CCR@FPR~$\uparrow$}& \multirow{2}{*}{ACC~$\uparrow$} \\
\cmidrule(lr){6-9}
& && && $10^{-4}$  & $10^{-3}$  & $10^{-2}$ & $10^{-1}$ \\
\midrule
\multirow{2}{*}{\begin{tabular}[c]{@{}c@{}}Texture \end{tabular}}&ResNet-18&0.89&99.42&99.67 / 99.05&54.03&75.02&87.81&94.62&95.40\\
&WideResNet-28&0.82&98.76&98.51 / 99.45& 53.66& 74.84& 87.96& 95.13& 96.28\\
\bottomrule
\end{tabular}
\vspace{-5pt}
\end{table*}

\begin{table}[t!]
    \caption{{Ablation study on GAT on the CIFAR-10  benchmarks. We report the mean values of all unlabeled datasets.}}
    \vspace{-5pt}
	\centering
 \small
	  \setlength{\tabcolsep}{9.5 pt}
	\begin{tabular}{c| c c c c c c c c c}
		\toprule
         {Modules}& {FPR95~$\downarrow$} & {AUROC~$\uparrow$ }& {AUPR(In/Out)~$\uparrow$}\\
        \hline
        {GNN}& {23.85}&{89.74}& {90.11 / 88.72} \\
       {Graphsage}& {16.08}& {86.07}& {87.10 / 86.09}\\
        {\textbf{GAT}}&{\textbf{7.90} }&{\textbf{97.42}}&{\textbf{97.45} / \textbf{96.86}}\\
        \bottomrule
	\end{tabular}
	\label{tab:gat}
 \vspace{-5pt}
\end{table}

\begin{figure}[t!]
    \centering
    \includegraphics[width=0.24\textwidth]{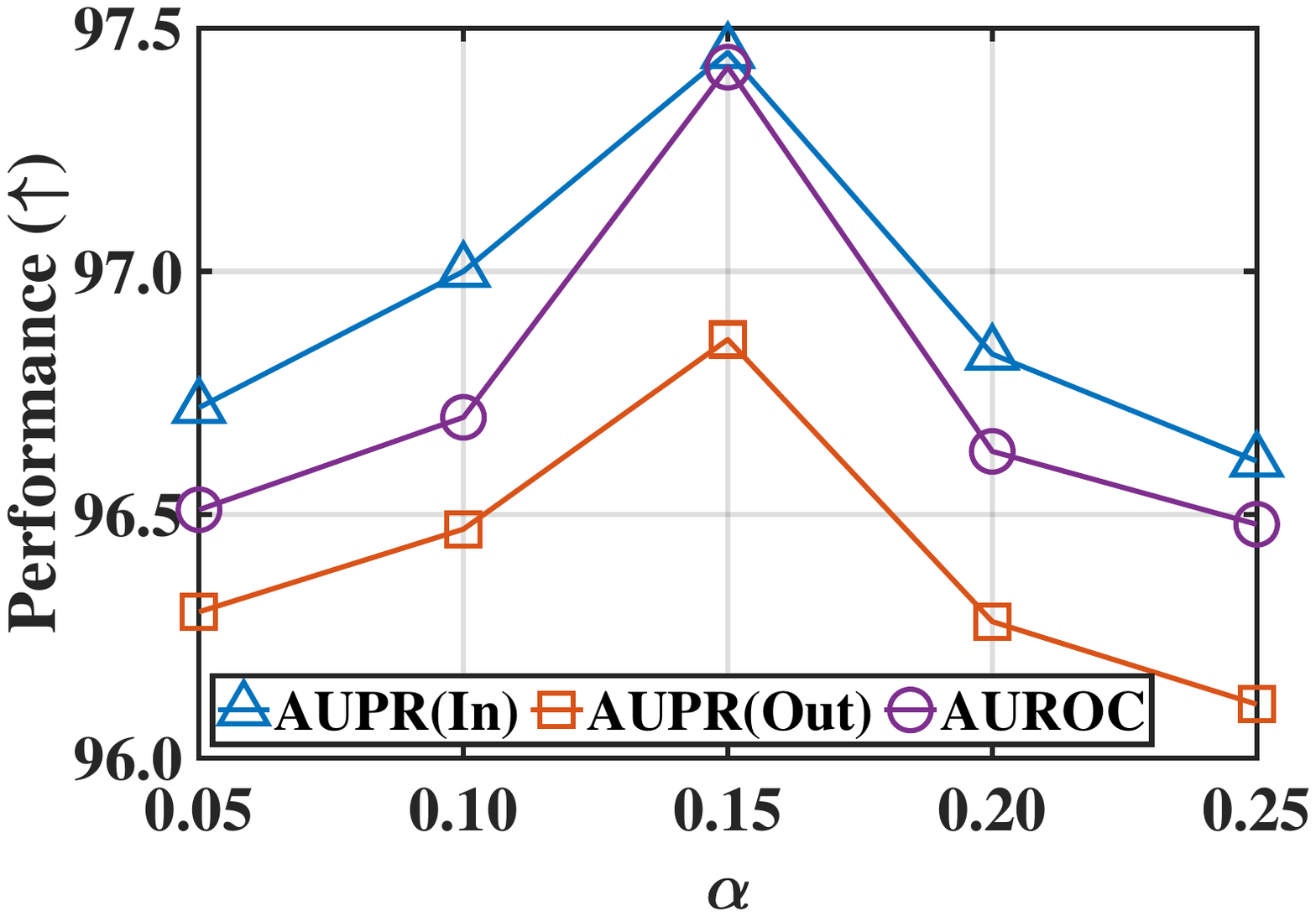}
    \hspace{-0.10in}
    \includegraphics[width=0.24\textwidth]{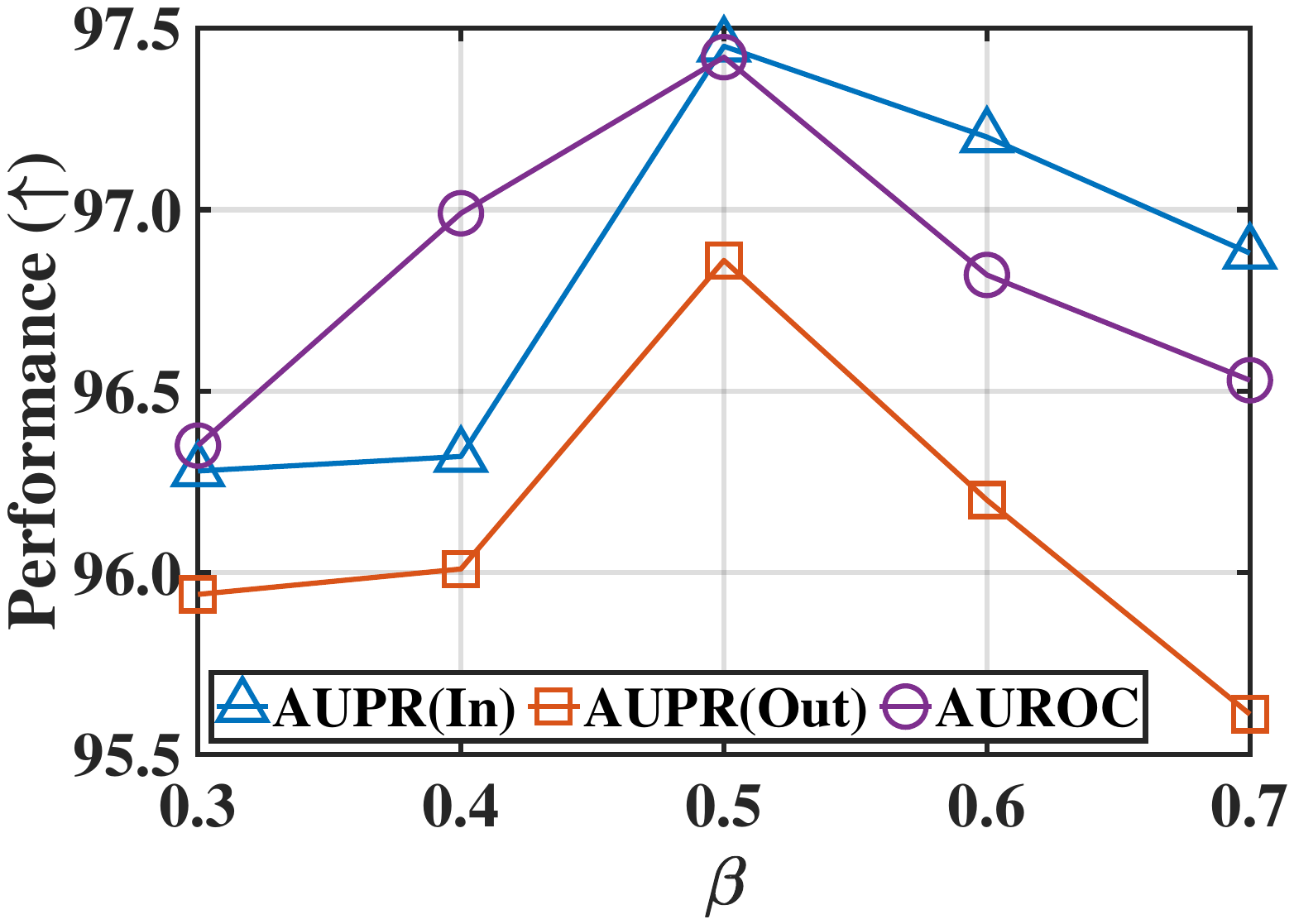}
     \hspace{-0.10in} 
    \includegraphics[width=0.24\textwidth]{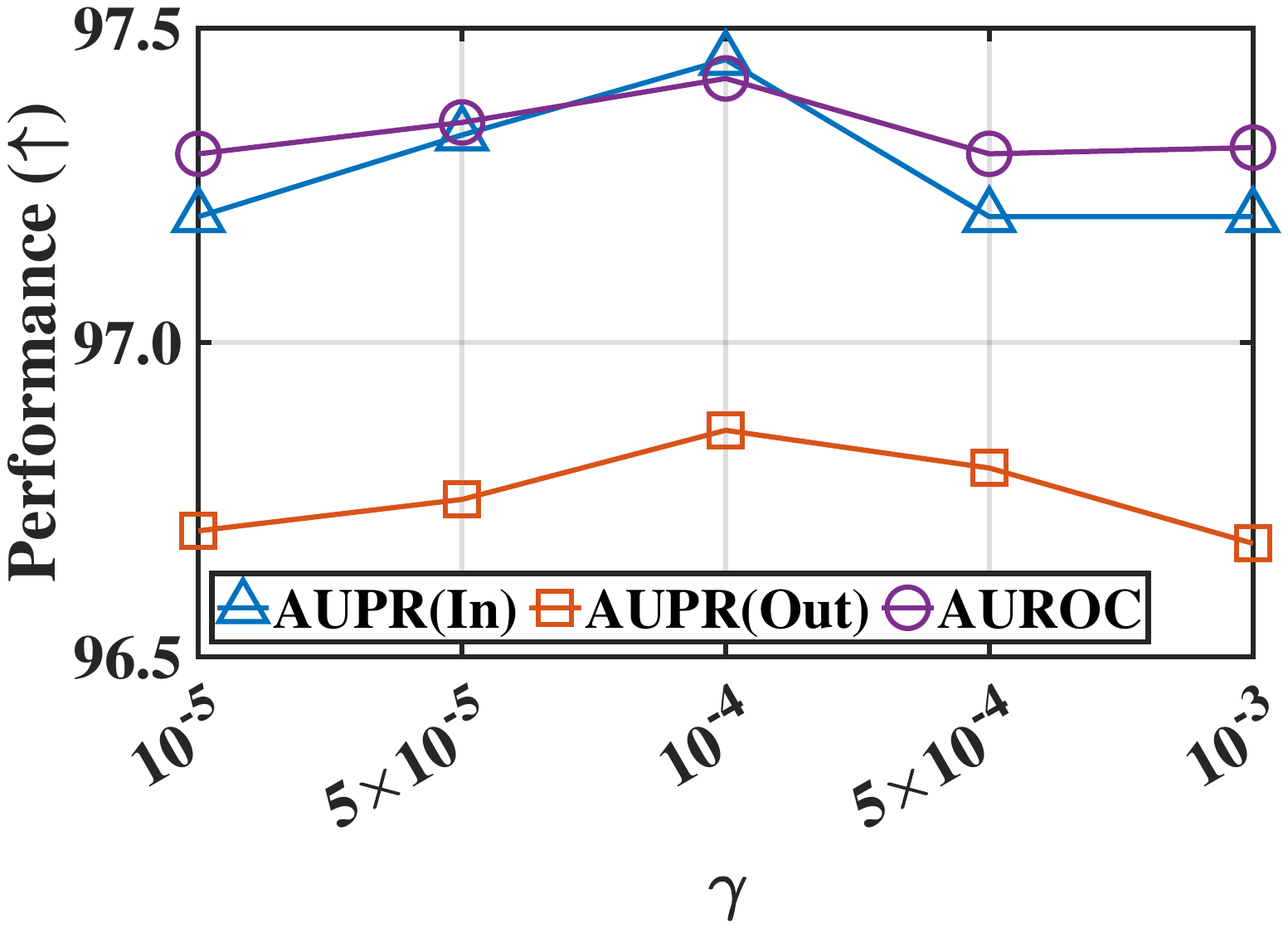}
    \hspace{-0.10in}
    \includegraphics[width=0.24\textwidth]{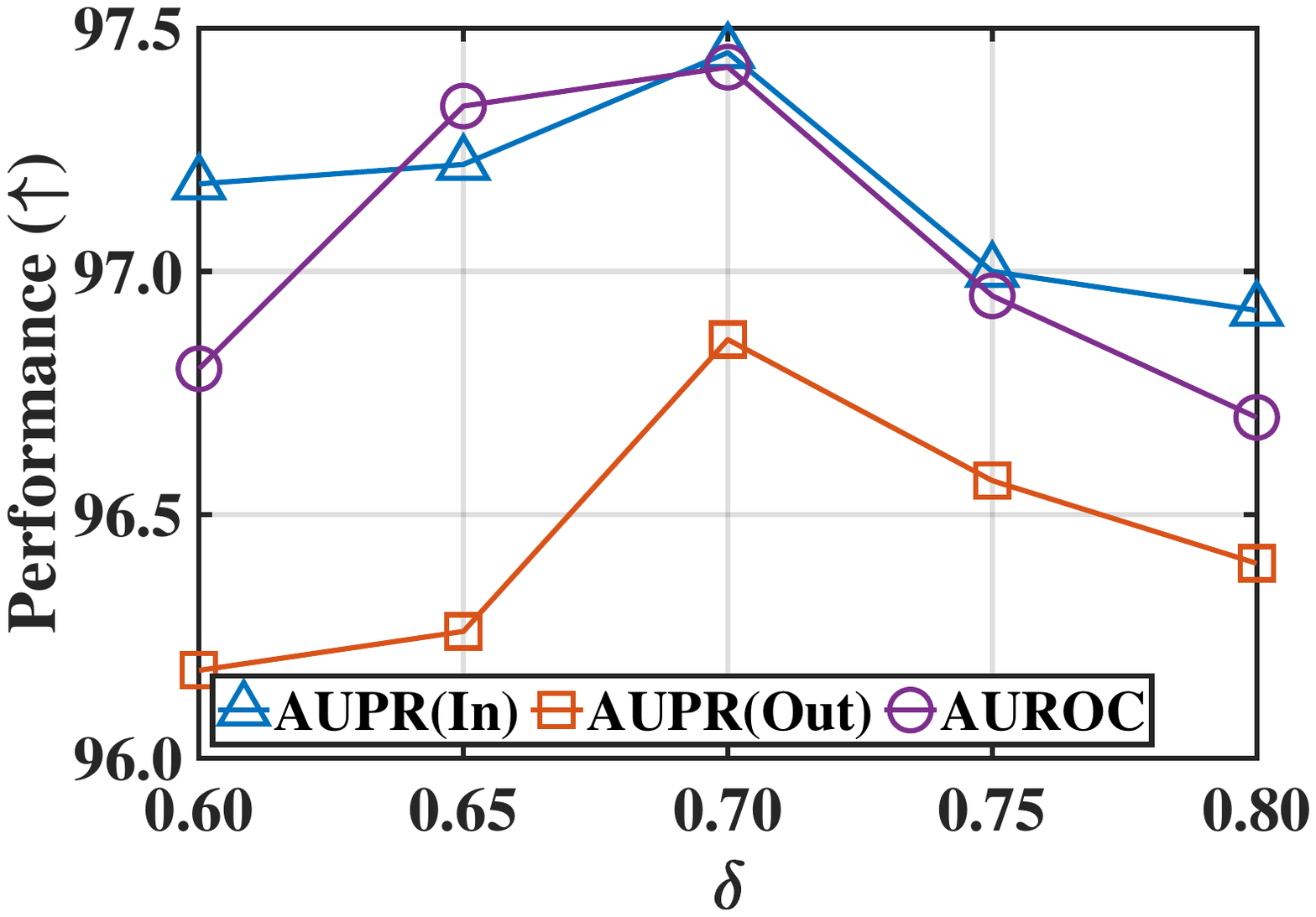}
    \hspace{-0.10in}
    \includegraphics[width=0.24\textwidth]{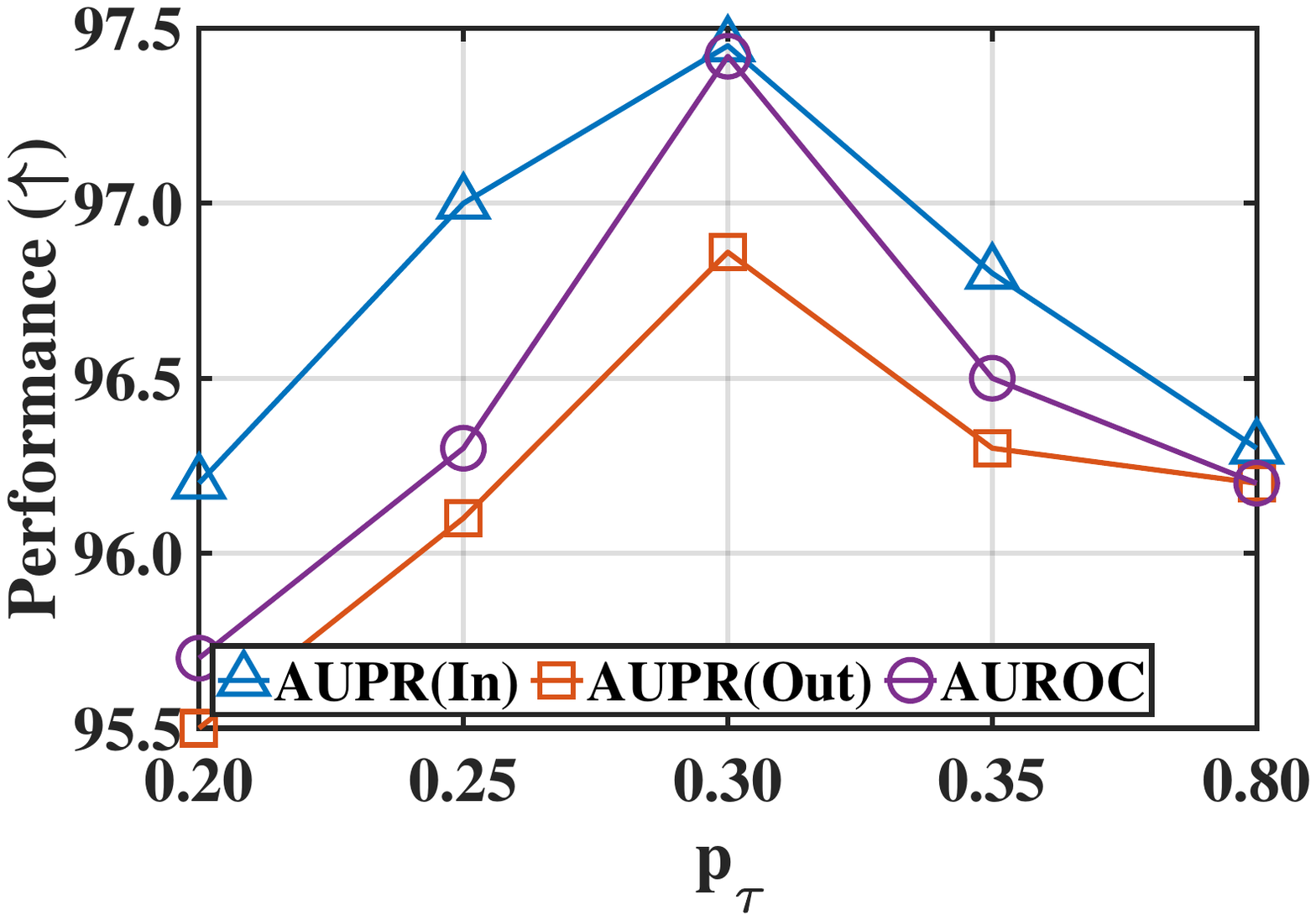}
    \hspace{-0.10in}
    \includegraphics[width=0.24\textwidth]{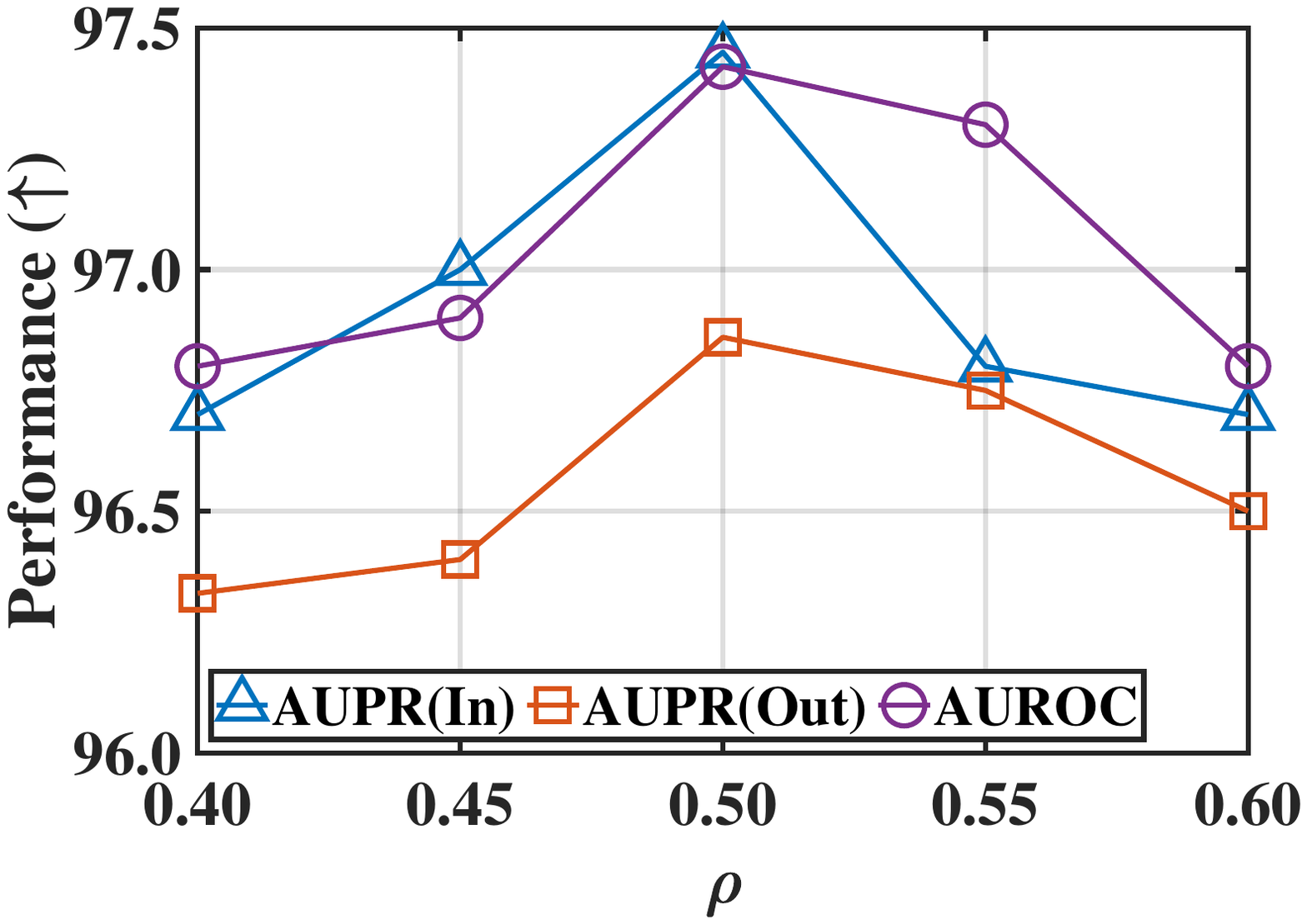}
    \hspace{-0.10in}
 	\caption{Effect of parameter on the CIFAR-10 benchmark.}
	\label{fig:para}
 \vspace{-5pt}
\end{figure}

\subsubsection{Effect of feature encoder} 
Following~\cite{yang2021semantically}, we try to utilize different image feature encoders (ResNet-18 and WideResNet-28) on the CIFAR-10 benchmark to further evaluate the performance of our AHGC. 
As shown in Table \ref{T:diff_feature_encoder}, we investigate different variants of feature encoders on all the tasks.  
Obviously, for the  WideResNet-28 network,   AHGC still obtains the satisfactory performance than state-of-the-art compared methods in Table \ref{T:CIFAR10_res18} and Table \ref{T:CIFAR100_res18} in most cases. 
The satisfactory performance of  AHGC shows its effectiveness  in both   OOD detection and  ID classification. An interesting finding is that WideResNet-28 achieves similar performance compared to ResNet-18. Since WideResNet-28 has a higher computational complexity, we choose ResNet-18 as the backbone.

{\subsubsection{Impact of graph attention network (GAT)} 
To explore our utilized  graph attention network, we conduct  a corresponding ablation study. We compare two popularly-used graph cut network: GNN \cite{scarselli2008graph} and Graphsage \cite{hamilton2017inductive}.
The ablation study results on GAT are reported in Table \ref{tab:gat}. Obviously, the GAT module is more effective than GNN and Graphsage on the OOD detection task.
}

\subsubsection{Influence of hyper-parameters ($\alpha$, $\beta$, $\gamma$, $\delta$, $p_\tau$ and $\rho$) }
We introduce some losses to supervise the training process. To evaluate the significance of these losses, we conduct  an ablation study to analyze hyper-parameters ($\alpha$, $\beta$, $\gamma$, $\delta$, $p_\tau$ and $\rho$). Fig. \ref{fig:para} shows the ablation study results. Obviously,
the optimal values for  these parameters are: $\alpha=0.15$, $\beta=0.5$, $\gamma=10^{-4}$, $\delta=0.7$, $p_\tau=0.3$ and $\rho=0.5$. Therefore, we utilize the above parameter setting in this paper.

\begin{figure}[t!]
\centering
\includegraphics[width=0.48\textwidth]{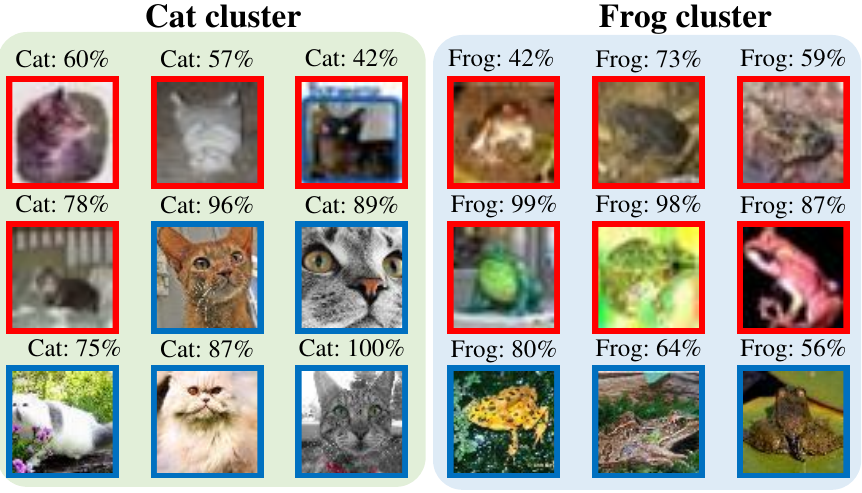}
\caption{\textbf{Visualization of the proposed AHGC on the CIFAR-10 benchmark.} We show partial samples from two subgraphs, where the labeled percentages of both ``Cat' and ``Frog'' clusters are larger than 85\%. Above each image, we report our predicted label and the  corresponding probability.
The labeled images with \textcolor{red}{red} edges are from the CIFAR-10 dataset. The unlabeled images with \textcolor{blue}{blue} edges are from the unlabeled Tiny-ImageNet dataset. The visualization results
illustrate that our proposed AHGC can assign correct labels to  unlabeled ID samples with high confidence. Best viewed in color.}
\vspace{-5pt}
\label{Fig:Visualization}
\end{figure}

\subsection{Qualitative Analysis}\label{subsection:qualitative}
To qualitatively investigate the effectiveness of our AHGC, we report some representative examples from CIFAR-10 benchmark in Fig. \ref{Fig:Visualization}. Obviously, our AHGC can assign the correct labels to unlabeled ID samples with high confidence, which verifies the effectiveness of our intra-subgraph label assignment module. Besides, no unlabeled OOD sample is mistakenly into ID cluster. It is because our attention-aware graph cut module can correctly explore the semantics relationship between unlabeled and labeled samples from multiple granularity  for final OOD detection.

\section{Conclusion}\label{conc}
In this paper, we propose a novel Adaptive Hierarchical Graph Cut network (AHGC)
to deeply explore the semantics relationship between labeled
and unlabeled samples with different granularity. Specifically, we first  explore the relationship between different images by constructing a hierarchical KNN graph based on the cosine similarity. By integrating the linkage and density information, we  conduct  graph cut on the graph. Then, we design an iterative labeling strategy to encourage semantic alignment between different samples by   a predefined threshold.   
Finally, we augment each image to further improve the model generalization.
Experimental results show the effectiveness of our AHGC. In representative cases, AHGC outperforms all state-of-the-art OOD detection methods by 81.24\% on the CIFAR-100 benchmark and by 40.47\% on the CIFAR-10 benchmark in terms of ``FPR95''.
In the future, 
we will extend  AHGC to  detect OOD object in  videos.

\bibliographystyle{IEEEtran}
\bibliography{main}

\end{document}